\documentclass[11pt]{article}

\usepackage[margin=1in]{geometry}
\usepackage{setspace}
\usepackage{microtype}
\usepackage[disable,textsize=tiny]{todonotes}

\usepackage[utf8]{inputenc}
\usepackage[T1]{fontenc}

\usepackage{amsmath, amsthm, amssymb, bbm, xparse, xspace} 
\usepackage{mathtools}
\usepackage{amsfonts}
\usepackage{nicefrac}

\usepackage{graphicx}
\usepackage{subcaption}
\usepackage{multirow}
\usepackage{booktabs}
\usepackage[table,dvipsnames]{xcolor}
\usepackage{colortbl}
\usepackage{float}
\usepackage[shortlabels]{enumitem}
\setlist[enumerate]{nosep, topsep=2pt}

\usepackage{algorithm}
\usepackage{algorithmic}

\usepackage[round]{natbib}

\usepackage{hyperref}
\usepackage{url}
\usepackage{xurl}
\usepackage[capitalize,noabbrev]{cleveref}
\crefname{subsection}{subsection}{subsections}

\usepackage{etoolbox}
\usepackage{comment}
\usepackage[english]{babel}
\usepackage[autostyle, english = american]{csquotes}
\MakeOuterQuote{"}

\theoremstyle{plain}
\newtheorem{theorem}{Theorem}[section]

\newtheorem{lemma}[theorem]{Lemma}

\theoremstyle{definition}

\theoremstyle{remark}

\NewDocumentEnvironment{myproof}{o}
  {\IfNoValueTF{#1}{\paragraph{{Proof.} }} {\paragraph{{#1.} }} }
  {\hfill$\qedsymbol$}

\newcommand{\argmax}{\mathrm{argmax}}

\newcommand{\bI}{\mathbbm{1}}
\newcommand{\bE}{\mathbb{E}}

\newcommand{\bR}{\mathbb{R}}

\newcommand{\hc}{{\hat{c}}}
\newcommand{\hn}{{\hat{n}}}

\newcommand{\hm}{{\widehat{m}}}

\newcommand{\hF}{{\widehat{F}}}
\newcommand{\hR}{{\widehat{R}}}
\newcommand{\ha}{{\hat{a}}}
\newcommand{\hxi}{{\hat{\xi}}}
\newcommand{\cA}{{\mathcal{A}}}
\newcommand{\CR}{{C}}
\newcommand{\supp}{\mathrm{supp}}

\newcommand{\worstCR}{\mathsf{WorstCR}}
\newcommand{\avgCR}{\mathsf{AvgCR}}
\newcommand{\Wasserstein}{\mathsf{Wasserstein}}
\newcommand{\Kolmogorov}{\mathsf{Kolmogorov}}

\usepackage[textsize=tiny]{todonotes}

\title{LLM-SAA: LLM-persona Generated Distributions for Decision-making\thanks{Code: \url{https://github.com/yunhanchen2/Evaluating-LLM-persona-Generated-Distributions-for-Decision-making}}}

\author{Jackie Baek\thanks{Stern School of Business, New York University, \href{mailto:baek@stern.nyu.edu}{baek@stern.nyu.edu}}
\and
Yunhan Chen\thanks{Department of Computer Science, Columbia University, \href{mailto:yc4512@columbia.edu}{yc4512@columbia.edu}}
\and
Ziyu Chi\thanks{Stern School of Business, New York University, \href{mailto:zc3725@stern.nyu.edu}{zc3725@stern.nyu.edu}}
\and
Will Ma\thanks{Graduate School of Business and Data Science Institute, Columbia University, \href{mailto:wm2428@gsb.columbia.edu}{wm2428@gsb.columbia.edu}}
}
\date{}

\begin{document}

\maketitle

\begin{abstract}
LLMs can generate a wealth of data, ranging from 
simulated personas imitating human valuations and preferences, to demand forecasts based on world knowledge.
But how well do such LLM-generated distributions support downstream decision-making?
For example, when pricing a new product, a firm could prompt an LLM to simulate how much consumers are willing to pay based on a product description, but how useful is the resulting distribution for optimizing the price?
We consider an approach where an LLM is used to construct an estimated distribution and the decision is then optimized under that distribution, referred to as LLM Sample Average Approximation (LLM-SAA).
We also propose decision-aware metrics to evaluate the performance of LLM-SAA.
Taking three canonical decision-making problems (assortment optimization, pricing, and newsvendor) as examples, we find that LLM-generated distributions are practically useful, especially in low-data regimes. We also show that decision-agnostic metrics such as Wasserstein distance can be misleading when evaluating these distributions for decision-making.
\end{abstract}

\section{Introduction}

In many problems, the optimal decision depends on a distribution of possible outcomes. Examples include deciding a single product assortment for a distribution of customer preferences, 
setting a price for a distribution of willingness-to-pay, or selecting an inventory level for a distribution of demand scenarios. Here, outcomes can vary either across a population (e.g., heterogeneous preferences) or over time (e.g., stochastic demand), and the true distribution $F$ governing these outcomes is typically unknown to the decision-maker, and difficult to characterize. While historical data is the canonical foundation for quantifying uncertainty about $F$, it is frequently sparse, costly to collect, or entirely unavailable for new products and evolving market conditions.

Large language models (LLMs) offer a new data source in precisely these low-data regimes. When real observations are limited or unavailable, an LLM can be prompted to generate synthetic samples or to describe a distribution using its knowledge of the world. For example, in a pricing problem, one can prompt an LLM to generate samples of willingness-to-pay given a product description and/or customer characteristics, which induces an empirical distribution that can be fed into a pricing optimizer. This suggests a simple pipeline which we call \textbf{LLM-SAA}: elicit an LLM-generated distribution $\hF$ for the outcome of interest, and then optimize the decision based on $\hF$.

This idea is related to a growing literature that uses LLMs to generate synthetic data by ``simulating humans'', for example to pilot social science experiments or generate survey samples to understand public opinion (see \cref{sec:simulacraRelWk}).
Our work uses LLMs similarly but for a new application: making downstream operational decisions. For this application, a central challenge is evaluation: how do we determine whether an LLM-generated distribution $\hF$ is ``good'' for decision-making when the true distribution is $F$? Prior work on LLM-based simulation has largely relied on \textit{decision-agnostic} notions of ``distributional alignment'', measuring statistical distances between the distributions $\hF$ and $F$.
For our application, however, statistical closeness is not the goal. What ultimately matters is decision quality: whether optimizing on $\hF$ yields actions that perform well under $F$.

This landscape motivates two connected research questions: 
\begin{enumerate} 
\item \textit{How should we evaluate LLM-generated distributions when the objective is downstream decision quality?} 
\item \textit{Can LLM-generated distributions be helpful for making downstream operational decisions?} 
\end{enumerate}

Answering these questions is subtle.
On one hand, $\hF$ can lead to the right decision for completely the wrong reason, yielding a ``lucky guess'' when $\hF$ had no resemblance to $F$.
On the other hand, expecting $\hF$ to look \textit{exactly} like $F$ may be unnecessary for systematically making good decisions.
We balance between these extremes 
by testing a \textit{range} of exogenous problem parameters that change the optimal decision under $F$.
This makes it harder to repeatedly make a ``lucky guess'', yet does not require $\hF$ to be identical to $F$.
As an example, for the pricing problem, we can adjust the cost of procuring the product, which affects its optimal price.
For $\hF$ to yield a good decision under a range of costs, 
it just needs to correctly capture the \textit{relevant} features of $F$, such as where purchase probabilities drop sharply; see \Cref{app:wtpPlots} for an illustration of this.

Continuing along our idea, we must define a reasonable range of problem parameters to evaluate.
To avoid arbitrariness, we resort to \textit{worst-case analysis}, where we identify the problem parameters that make the decision suggested by $\hF$ look pessimal relative to the optimal decision knowing $F$.
To our knowledge, this is a new concept in data-driven and robust optimization (see \Cref{sec:optRelWk}); computing these worst-case parameters for a given $\hF$ and $F$ poses an interesting problem.

\paragraph{Contributions and approach.}
Our main contribution is to consider the LLM-SAA pipeline and propose evaluation metrics for it, whereby we measure downstream decision quality rather than the statistical similarity of the LLM-generated distributions. We apply this methodology to three well-studied problems in operations research: assortment, pricing, and newsvendor.

We define decision-aware performance metrics based on \emph{competitive ratios} that compare the performance of the action optimized under $\hF$ to the true optimal action under $F$. 
We study both the worst-case competitive ratio over problem parameters, and  an average-case competitive ratio over a parameter distribution.
Our main technical contribution is showing how to compute the worst-case competitive ratio under the three settings.
Moreover, we show that this worst-case metric is reasonably concordant with average-case parameter ranges that one may deem relevant \citep[cf.][]{roughgarden2021beyond}.

We consider multiple \textit{generation methods} for eliciting the LLM distribution $\hF$; these methods differ in what information is provided to the LLM and how it is prompted. 
We leverage real-world datasets across all three problem settings to evaluate LLM-generated distributions against baselines such as uniform distributions and $d$-sample empirical distributions.
We evaluate performance using the worst-case and average-case competitive ratios, and also contrast these with standard decision-agnostic distances such as Wasserstein and Kolmogorov distances.
We summarize our main takeaways:
\begin{enumerate}[I.]
\item  LLM-SAA can be practically useful in low-data regimes: it consistently outperforms uninformed baselines and can be competitive with empirical distributions on a small number of real samples;
\item  Decision-agnostic distances can be misleading in measuring whether a generated distribution is good for decision-making;
\item Steering with personas can improve decision-making, even though the persona-based LLM responses do not accurately match true responses at an individual level.
\end{enumerate}

\section{Related Literature}

\paragraph{Using LLMs to Simulate Humans.}\label{sec:simulacraRelWk}

A growing literature proposes using LLMs as ``synthetic respondents'' to approximate human survey responses and choice behavior in social science and market research \citep{aher2023using, argyle2023out, brand2023using,horton2023large,park2023generative, manning2024automated, wang2024large}.
Despite their promise, some papers caution against using LLMs as direct human replacements, citing concerns about representativeness and behavioral fidelity \citep{dillion2023can,sarstedt2024using,gao2025take}.
Consequently, a substantial line of recent work focuses on benchmarking and enhancing LLM-based human simulation
\citep{samuel2024personagym, xie2024humansimulacra, qu2024performance, meister2025benchmarking, suh2025language, toubia2025database}.
While these existing benchmarks assess fidelity at the individual or aggregate level, our work introduces decision-aware evaluation metrics to assess the utility of LLM-generated population distributions for downstream decision-making.

\paragraph{Data-driven and Robust Optimization.} \label{sec:optRelWk}

We study stochastic optimization problems where the distribution $F$ is unknown.
When historical samples are available,
a standard approach is Sample Average Approximation (SAA), replacing expectations under $F$ with empirical averages \citep{shapiro2014}. In the absence of data, one typically resorts to robust optimization \cite{bental2009} or online algorithms analyzed via competitive ratios \cite{borodin1998}. 
In the age of LLMs, however, SAA can be used even in no-data regimes. We refer to this approach as \emph{LLM-SAA}: we use an LLM to construct an estimated distribution $\widehat{F}$ (often via synthetic samples), and then optimize the downstream decision under $\widehat{F}$ via SAA.
We study how different generation methods (prompting strategies and LLM models) affect $\widehat{F}$. For evaluation, we consider worst-case competitive ratios over known instance parameters $\theta$ while holding $F$ and $\widehat{F}$ fixed, which is different from typical analyses that consider worst cases over unknowns in $F$.



\paragraph{LLMs for Operational Decisions.}
\label{sec:llm_or_decisions}
There is a growing literature and application for LLMs inside decision-making pipelines.
Work that evaluates LLMs directly as decision makers in canonical OR tasks finds that plausible language outputs can still correspond to biased or unreliable actions, for example in the newsvendor problem \citep{liu2025large}. However, recent work suggests that LLMs can straight-up outperform humans in the challenging beer game for inventory management \citep{LongSimchiLeviCalmonCalmon2025Autonomous}.
An extensive study using an LLM to directly manage a real vending machine is described in \citet{backlund2025vendingbench}.
Related to pricing, \citet{cao2026llm} study the potentially collusive consequences of using LLMs to make pricing decisions.

A complementary set of papers argues that the central opportunity is to pair LLMs with optimization to reduce modeling and communication frictions.
\citet{wang2025omgpt} use a transformer (the architecture behind LLMs) to predict optimal decisions.
\citet{duan2025ask} propose a hybrid approach in which an LLM elicits and clarifies inputs and then a solver computes the decision, reporting large cost improvements over an end-to-end LLM baseline.
A line of work explicitly uses LLMs to formulate textbook optimization \citep{huang2025orlm} and dynamic programming \citep{zhou2025auto} problems, feeding them into optimization solvers.
Broader discussion positions LLMs as a planning layer around optimization tools \citep{simchi2025large}, describing research agendas for AI in supply chains \citep{cohen2025supply}.

Our work fits into the latter ``LLM plus optimization'' literature, but in contrast, focuses on \textit{using LLMs to generate data for optimization problems} under uncertainty, i.e., LLM-SAA.  Evaluating the generated distributions wholistically is subtle, which is the main investigation of our work.

\section{Problems, Methods, and Metrics}

We consider problems where there is an unknown distribution $F$ over outcomes $\xi$ (e.g., $\xi$ is a customer preference ranking).
There is a set of known parameters $\theta$, which dictate the feasible action set $\cA_\theta$ and the 
reward $r_\theta(a,\xi)$ that is earned when action $a \in \cA_\theta$ (e.g., $a$ is an assortment of products) is taken under realized outcome $\xi$. 
The objective is to decide an action to maximize the expected reward $R_\theta(a):=\bE_{\xi\sim F}[r_\theta(a,\xi)]$,
where $a^* \in \argmax_{a \in \cA_\theta} R_\theta(a)$ denotes an optimal action.
Our overall approach is LLM-SAA: we use an LLM to construct an estimated distribution $\hF$, then choose the action $\ha$ by solving the corresponding sample-average approximation under $\hF$.
Specifically, letting $\hR_\theta(a):=\bE_{\xi\sim \hF}[r_\theta(a,\xi)]$ be the estimated expected rewards, the action is chosen as $\ha\in\argmax_{a\in\cA_\theta}\hR_\theta(a)$.

We consider  three well-studied problems in decision-making under uncertainty.
\begin{itemize}
\item  \textbf{Assortment.}\footnote{Standard assortment optimization focuses on revenue maximization \citep{talluri2004revenue}.
We instead study utility maximization like \citet{sumida2021revenueutility}, but define it for the ranking-based choice model \citep{farias2013nonparametric}.
}
There are $n$ items, indexed by $[n]:=\{1,\dots,n\}$. An outcome $\xi$ is a strict ranking of the items, written $\xi=(\xi_1,\dots,\xi_n)$, where $\xi_1$ is the most-preferred item and $\xi_n$ is least-preferred. An action $a \subseteq [n]$ is the set of offered items,  where $a$ must lie in some feasible family $\cA_\theta$ (e.g., $\cA_\theta=\{a:|a|\le k\}$, where $k$ is a size constraint).
Given assortment $a$, a customer with ranking $\xi$ derives highest utility if their most-preferred item $\xi_1$ is found in $a$, 2nd-highest utility if $\xi_1\notin a$ but $\xi_2\in a$, and so forth.  Formally,
\[
r_\theta(a,\xi)=r_{\min\{p\in[n]: \xi_p\in a\}}
\]
where $(r_p)_{p\in[n]}$ denotes the \textit{position-dependent rewards} satisfying 
$r_1\ge \cdots \ge r_n \ge 0$. The parameters $\theta$ consist of the reward vector $(r_p)_{p\in[n]}$ and any parameters, e.g.~$k$, that define the feasible family $\cA_\theta$.

\item \textbf{Pricing.}
The outcome $\xi\ge 0$ denotes a buyer's willingness-to-pay for a product. The action is a price $a\ge 0$ for the product. 
The parameter $\theta = c\ge 0$ defines the item's unit cost, and the reward is the profit from selling the item,
\[
r_\theta(a,\xi) = (a-c)\mathbf{1}\{\xi \ge a\}.
\]

\item 
\textbf{Newsvendor.}
The outcome $\xi\ge 0$ denotes the realized demand for an item over a selling period.
The action is an inventory level $a\ge 0$. 
The parameter $\theta = q\in(0,1)$ defines a reward function
\[
r_\theta(a,\xi)= -\left(q[\xi-a]^+ + (1-q)[a-\xi]^+\right),
\]
where $[z]^+:=\max\{z,0\}$ for all $z\in\bR$.
Here, the reward is the negative of a \textit{newsvendor loss function} in which the unit cost of understocking inventory (leading to $[\xi-a]^+>0$) is $q$ and the unit cost of overstocking inventory (leading to $[a-\xi]^+>0$) is $1-q$.
\end{itemize}
For each problem, we define the ground-truth distribution $F$ based on real dataset(s).
The optimal action $a^*$ depends on both the known instance parameters $\theta$ and the unknown $F$.

\subsection{Generation Methods and Baselines} \label{sec:generation_methods}

We estimate the unknown $F$ using a distribution $\hF$ generated by an LLM,
which can be prompted in different ways and given varying levels of information.
We compare against baselines that are given similar levels of information.
The exact prompts and baselines are problem-dependent and deferred to \Cref{sec:assortment,sec:pricing,sec:newsvendor} for assortment, pricing, and newsvendor respectively.
Here we only give a high-level taxonomy of the different prompting methods and baselines, noting that we capture all of the prompting methods described in \citet[\S3.1--3.2]{meister2025benchmarking}.

In terms of information, we always give the LLM background about where the ground-truth distribution $F$ came from, including description of product(s) and geographical location of customers.
For the pricing and newsvendor problems which have real-valued outcomes, we give the LLM numerical values to anchor on, but give the same anchors to the random baseline that is compared against.
In the \textbf{Few-shot} setting, the LLM is given additional information in the form of examples, either of outcomes similar to those in $F$, or of distributions similar to $F$.

Orthogonal to the information setting, we test the following prompting methods.
\begin{enumerate}
\item \textbf{Sampling.} The LLM is prompted to generate a single outcome $\xi$ many times independently, and then these $\xi$'s are combined to form the estimated distribution $\hF$.

\item \textbf{Persona-sampling.} Same as sampling, but the LLM is prompted to imitate a different customer ``persona'' each time (e.g.\ if the persona is ``rich person'', then the LLM might generate a high willingness-to-pay $\xi$).
We use a collection of personas that resembles, but is not identical to, the true characteristics of customers that generated the ground-truth distribution $F$.

\item \textbf{Batch-generation.} The LLM is prompted to generate a batch of outcomes $\xi$ using a single query; $\hF$ is the uniform distribution over this batch of outcomes.

\item \textbf{Description.} The LLM is asked to describe the distribution $\hF$, where the prompt is problem-dependent.
\end{enumerate}

We compare against non-LLM baselines that may utilize additional information:

\begin{itemize}
\item \textbf{Random.} A distribution that consists of uniformly random outcomes, drawn over a reasonably-anchored numerical range if the outcome is real-valued.
\item \textbf{$d$ real data.} A distribution consisting of $d$ samples drawn IID from the true distribution $F$.
\end{itemize}

We re-iterate that our focus is on regimes with low or no data, e.g.\ assortment optimization, pricing, and Newsvendor for new products or markets.  These are commonly-encountered business situations 
\citep{baardman2018leveraging, allouah2023optimal}, in which pre-LLM work had to resort to decisions that were uninformed of the distribution 
\citep{perakis2008regret, cohen2021simple}
We also note that in these problems, even the small values of $d$ we use provides a surprisingly strong baseline 
\citep{besbes2023big}
For the assortment problem, we also compare against a baseline that gives each item a univariate score (see \Cref{sec:distMatters}).

\subsection{Metrics} \label{sec:measures}

There are different ways of measuring how far off the estimated distribution $\hF$ is from the true distribution $F$.
We consider the following (asymmetric) distance metrics between $F$ and $\hF$.

\paragraph{Decision-aware metrics, via competitive ratios.}
Given the LLM-SAA decision $\ha$, we ultimately care about $R_\theta(\ha)$, the expected reward of our decision $\ha$ on the true distribution $F$.
We compare $R_\theta(\ha)$ against $R_\theta(a^*)$, which is the expected reward of an optimal action $a^*$ knowing the ground-truth distribution $F$, by measuring the \textit{competitive ratio}
\begin{align} \label{eqn:crObj}
\CR_\theta(F,\hF) \;:=\; \frac{\min\{|R_\theta(a^*)|,|R_\theta(\ha)|\}}{\max\{|R_\theta(a^*)|,|R_\theta(\ha)|\}}.
\end{align}
For problems with positive reward functions (assortment, pricing), we have $\CR_\theta(F,\hF)=R_\theta(\ha)/R_\theta(a^*)$, while for problems with negative rewards (newsvendor), we have $\CR_\theta(F,\hF)=R_\theta(a^*)/R_\theta(\ha)$.
In both cases, $\CR_\theta(F,\hF)\le1$, with bigger competitive ratios being better and indicating that optimizing on $\hF$ leads to a good decision on $F$.

Given $F$ and $\hF$, the evaluation $\CR_\theta(F,\hF)$ still depends on the instance parameters $\theta$.  We measure both worst-case and average-case competitive ratios over $\theta$, where $\Theta$ denotes the parameter space and $\Pi$ denotes a distribution over $\Theta$:
\begin{align*}
\worstCR(F,\hF) \; &:=\; \inf_{\theta\in\Theta} \CR_\theta(F,\hF), \\
\avgCR(F,\hF) \; &:=\; \bE_{\theta\sim \Pi}\!\left[\CR_\theta(F,\hF)\right].
\end{align*}

Measuring $\worstCR(F,\hF)$ is difficult in general, but we show how to compute it for our three specific problems of interest---assortment, pricing, and newsvendor.  We formulate the problem as minimizing~\eqref{eqn:crObj} subject to
\begin{align}
a^* \in \argmax_{a \in \cA_\theta} R_\theta(a);\ \ha\in\argmax_{a\in\cA_\theta}\hR_\theta(a), \label{eqn:optimalityConds}
\end{align}
where $\theta\in\Theta$ and $a^*,\ha\in\cA_\theta$ are treated as decision variables while the optimality conditions~\eqref{eqn:optimalityConds} are treated as constraints.
Under this formulation of $\worstCR(F,\hF)$, tiebreaking favors the adversary, as they can select the $\ha\in\argmax_{a\in\cA_\theta}\hR_\theta(a)$ with lowest value of $R_\theta(\ha)$.  By contrast, when measuring $\avgCR(F,\hF)$, we fix an arbitrary rule for selecting $\ha$ given problem parameters $\theta$.

\paragraph{Decision-agnostic metrics.}
We also report standard distributional distances between $F$ and $\hF$ that do not depend on the decision problem. We consider the Wasserstein distance:
$
\Wasserstein(F,\hF)
\;:=\;
\inf_{G\in \Gamma(F,\hF)} \bE_{(\xi,\hxi)\sim G}\big[\|\xi-\hxi\|\big],
$
where $\Gamma(F,\hF)$ denotes the set of couplings with marginals $F$ and $\hF$.
For pricing and newsvendor, we let $\|\xi-\hxi\|:=|\xi-\hxi|$, and for assortment, we let $\|\xi-\hxi\|$ denote the Kendall-tau distance between rankings.

For pricing and newsvendor which have real-valued outcomes, we define $F(z):=\Pr_{\xi\sim F}[\xi\le z]$ to be the CDF, and additionally report the Kolmogorov distance:
$\Kolmogorov(F,\hF) \;:=\; \sup_{\xi\in\bR}\, \big|F(\xi)-\hF(\xi)\big|.
$
Finally, for the assortment and pricing problems, we also consider a metric for how well the LLM is able to imitate specific customer personas (see \Cref{sec:asstPersona,sec:pricingPersona}).

\section{Assortment problem: Details and Results} \label{sec:assortment}

\paragraph{Dataset.}
We use the \emph{SUSHI Preference Data Sets, Dataset A} \citep{kamishima_sushi}, collected via a questionnaire survey of respondents in Japan. Each respondent provides a strict ranking over $n=10$ sushi items.
We consider a single ground-truth $F$ defined by taking the uniform distribution over the first $600$ respondents' rankings.

\paragraph{Details of LLM generation methods and baselines.}
As background information, we always give the LLM a description of each sushi, and tell it that the respondents are from Japan.
In the \textbf{Few-shot} setting, we additionally give the LLM 6 randomly-drawn rankings of respondents \textit{outside} the first 600 (i.e., not lying in $F$).
Orthogonally, we consider all 4 ways of prompting described in \Cref{sec:generation_methods}, noting that \textbf{Persona-sampling} gives the LLM the gender, age group, current residence, and childhood residence of a respondent outside $F$; \textbf{Batch-generation} uses a batch size of 30; and \textbf{Description} asks the LLM to generate a score for each sushi, based on which it generates $\hF$ using a Plackett-Luce model \citep{Plackett1975}.
Exact prompts 
are provided in \cref{app:promptAsst}.

Here, the \textbf{Random} baseline is composed of 600 rankings drawn uniformly among all $n!$ permutations.
We also compare against a baseline that computes a univariate \textbf{Popularity score} for each item based on the true $F$. Specifically, for each sushi $j\in[n]$ we define
\begin{align} \label{eq:popularity}
S_j:=R_\theta(\{j\})=\bE_{\xi\sim F}\left[\sum_{p=1}^n r_p\bI(\xi_p=j)\right],
\end{align}
which is the expected reward earned from offering only $j$.
The baseline outputs the feasible assortment $a\in\cA_\theta$ that maximizes the total score $\sum_{j\in a}S_j$, which is quite a strong baseline in that it is based on the ground truth $F$.

\paragraph{Further problem details.}
We consider feasible families $\cA_\theta$ defined by ``knapsack'' constraints, where each item $j\in[n]$ has a size $s_j\ge 0$, and feasible assortments have total size bound by some budget.  We express the budget as a multiple $B$ of the average size $\frac1n\sum_{j=1}^n s_j$, so formally, we have
$\cA_\theta = \{ a\subseteq[n]: \sum_{j\in a} s_j \le B\frac1n\sum_{j=1}^n s_j\}$.
We interpret $B$ as the average number of items that would fit in the knapsack.

Unless otherwise stated, we use position-dependent rewards $r_p=10/p$ for all $p\in[n]$.
In \Cref{app:asstTables}, we additionally report results for alternative reward sequences $(r_p)_{p\in[n]}$; across those tables, the only change is the choice of $(r_p)$.

For $\avgCR(F,\hF)$, we consider three regimes of how the sizes $(s_j)_{j\in[n]}$ are drawn, which induce different distributions $\Pi$ over the parameters:
\textbf{Unit}: $s_j=1$ $\forall j\in[n]$;
\textbf{Random}: $s_j$ is drawn independently and uniformly from $[0.8,1.2]$ $\forall j\in[n]$, repeating 100 times so that $\Pi$ is a distribution over 100 random instances;
\textbf{Hard}: $s_j = S_j$ $\forall j\in[n]$, where $S_j$ is the popularity score defined in \eqref{eq:popularity}.

For each of the above regimes, we use two different budgets $B \in \{2, 5\}$.
The third regime is hard because the size of an item is exactly equal to its popularity score; hence a simple strategy of including popular items may not perform well.

\paragraph{Theoretical result.}
We show how to efficiently compute $\worstCR(F,\hF)=\inf_{(s_j)_{j\in[n]},B}R_\theta(\ha)/R_\theta(a^*)$, for any fixed position-dependent rewards $(r_p)_{p\in[n]}$.
Note that the functions $R_\theta(\cdot),\hR_\theta(\cdot)$ are fully fixed based on $(r_p)_{p\in[n]}$, so we can sort
the ratios $R_\theta(\ha)/R_\theta(a^*)$ over all pairs $a^*,\ha\subseteq[n]$, and find the pair with the smallest ratio for which there exists $\cA_\theta$ (constructed from $(s_j)_{j\in[n]},B$) such that~\eqref{eqn:optimalityConds} holds.
Clearly, a pre-requisite to~\eqref{eqn:optimalityConds} holding is that
\begin{align} \label{eqn:theoryPrereq}
R_\theta(a^*)\ge R_\theta(\ha);\ \hR_\theta(\ha)\ge\hR_\theta(a^*).
\end{align}

\begin{lemma}[proven in \Cref{pf:asstKey}] \label{lem:asstKey}
Let $a^*,\ha\subseteq[n]$ be assortments for which~\eqref{eqn:theoryPrereq} holds.
If $\min\{|a^*\setminus\ha|,|\ha\setminus a^*|\}\le 1$, then there exist sizes $(s_j)_{j\in[n]}$ and a budget $B$ such that only the sets $a^*,\ha$ can be maximal in $\cA_\theta$, and hence~\eqref{eqn:optimalityConds} can also hold.
Otherwise, $\cA_\theta$ must contain maximal sets other than $a^*,\ha$.
\end{lemma}

\Cref{lem:asstKey} implies that for any $a^*,\ha$ satisfying~\eqref{eqn:theoryPrereq} and $\min\{|a^*\setminus\ha|,|\ha\setminus a^*|\}\le 1$,
we can immediately show the existence of $\cA_\theta$ for which~\eqref{eqn:optimalityConds} holds,
because $\max_{a\in\cA_\theta}$ is always achieved at a maximal set (given that $R_\theta,\hR_\theta$ are monotonically increasing in $a$).

Equipped with \Cref{lem:asstKey}, our algorithm iterates\footnote{Our algorithm requires enumerating all $2^n$ assortments, leading to exponential runtime (reasonable in our setting with $n=10$).  Without our algorithm, enumerating all possible feasible families $\cA_\theta$ would take doubly-exponential runtime.} over $a^*,\ha\subseteq[n]$.  If~\eqref{eqn:theoryPrereq} holds and $R_\theta(\ha)/R_\theta(a^*)$ undershoots the smallest ratio found so far, both of which are easily checkable, then we see if there exists $\cA_\theta$ for which~\eqref{eqn:optimalityConds} holds.  By \Cref{lem:asstKey}, the answer is immediately Yes if $\min\{|a^*\setminus\ha|,|\ha\setminus a^*|\}\le 1$.  Otherwise, we show how to write a Linear Program (LP) to check if~\eqref{eqn:optimalityConds} can hold. 
Details are found in
\Cref{app:worstcrAsst}, where we also present some tricks to prune the search over $a^*,\ha\subseteq[n]$.

\subsection{Assortment problem: Empirical Results} \label{sec:asstEmpRes}

We compare 8 generation methods (2 information settings $\times$ 4 ways of prompting) across 4 LLM models (GPT-4o, GPT-5-mini, Gemini 3 Flash, Mistral Large 3), and 5 baseline methods (Random, 3 real data sizes, Popularity score), for generating an estimated distribution $\hF$.  We note that some methods are random and produce a different $\hF$ each time, so we report averages over multiple $\hF$'s.
Full generation details and results are deferred to \Cref{app:asstTables}.

\begin{figure*}[h]
\centering
        \includegraphics[width=0.7\textwidth]{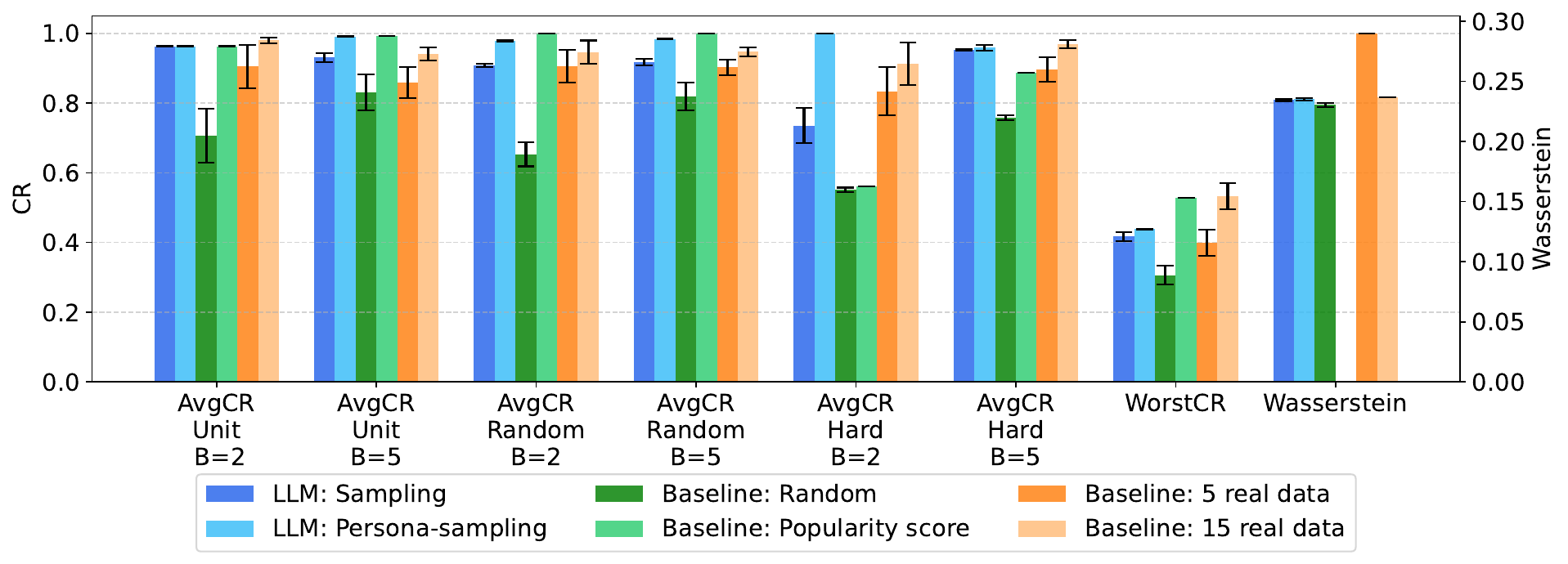}
    \caption{
    Assortment results, displaying means across 20 generations, with 95\% confidence intervals around the mean.  \textbf{Higher is better} for the CR metrics while \textbf{lower is better} for the Wasserstein metric. The LLM is GPT-4o.
    }
    \label{fig:assortment_bar}
\end{figure*}

\Cref{fig:assortment_bar} summarizes the findings, where we compare representative LLM generation methods to baselines, across decision-aware and decision-agnostic metrics.
We see that the LLM-based methods consistently outperform the Random baseline under the decision-aware competitive ratio (CR) metrics, indicating that the LLM's generated rankings based on world knowledge are definitely better than a random guess, when it comes to making assortment decisions.
This pattern robustly extends to other generation methods and LLM models (see \Cref{tab:assortment_results_10/i} for full results), noting that additional information (personas, few-shot examples) is not necessary for the LLM to beat the Random baseline.
Meanwhile, 15 real data provide a strong baseline, beatable only by select LLM methods.

We remark that $\worstCR$ provides a fairly concordant ``ranking'' of methods in terms of how well they would perform across different instance parameters, as captured in the $\avgCR$ metrics.
On the other hand,
the decision-agnostic Wasserstein metric paints a very different picture:
the Random baseline actually has the best (smallest) Wasserstein distance!
This highlights a central takeaway of our paper: standard distributional distances can be a very poor proxy for downstream decision quality in operational problems.


\paragraph{Popularity score can be insufficient.}\label{sec:distMatters}
The Popularity score baseline based on
the ground-truth $F$ is generally hard to beat, but as we see from \Cref{fig:assortment_bar}, it performs poorly in the Hard regime, where each item’s size exactly equals its popularity score.
In this regime, it is not sufficient to simply select popular items, since popular items are the largest and hence penalized by the budget constraint.

LLM-based methods outperform the Popularity score in this regime, suggesting they rely on more than just overall sushi popularity (even if their popularity estimates are imperfect). For example, LLMs can capture affinities---people who like \(j\) often also like \(j'\), so adding \(j\) has low marginal value when \(j'\) is already offered---and recognize ``polarizing'' sushis that tend to rank either first or last.


\label{sec:asstPersona}
\paragraph{Is the LLM able to imitate specific personas?}
In Persona-sampling, we prompt the LLM to generate a ranking by conditioning on an individual’s observed features, like in the literature discussed in \Cref{sec:simulacraRelWk}. 
We assess whether the LLM truly personalizes its rankings by comparing predicted rankings $\hat{\xi}^i$ to the true rankings $\xi^i$ for each individual $i$.
To quantify this, we compute two metrics:
\begin{itemize}
    \item $\mathsf{PersonaMAE}(F,\hF):=\frac{1}{m}\sum_{i=1}^m \|\xi^i-\hat{\xi}^{i}\|$: the average distance between the LLM’s prediction for a persona and that specific individual's true preference;
    \item $\mathsf{ShuffledMAE}(F,\hF):=\frac{1}{m^2}\sum_{i=1}^m \sum_{i'=1}^m \|\xi^i-\hat{\xi}^{i'}\|$: the average distance between the LLM’s prediction for a persona and the preference of a \textit{randomly} selected individual from the population.
\end{itemize}

If the LLM accurately tailors rankings to each individual, we would expect $\mathsf{PersonaMAE}$ to be substantially smaller than $\mathsf{ShuffledMAE}$. 
Instead, we find that these metrics are essentially identical: \textbf{0.367} for $\mathsf{PersonaMAE}$ and \textbf{0.369} for $\mathsf{ShuffledMAE}$.
(For comparison, the $\mathsf{ShuffledMAE}$ of the Random baseline is \textbf{0.497}.)

Despite this failure at the individual level, Persona-sampling achieves the highest CR metrics. 
These results imply that high-fidelity ``human simulation'' is not a prerequisite to adding value for downstream decision-making.
While existing literature often focuses on the accuracy of the individual imitation, our results show that persona-steering can be valuable because it encourages the LLM to represent the broader diversity and range of the population distribution.



\section{Pricing problem: Details and Results} \label{sec:pricing}

\begin{figure*}[h]
\centering
        \includegraphics[width=0.7\textwidth]{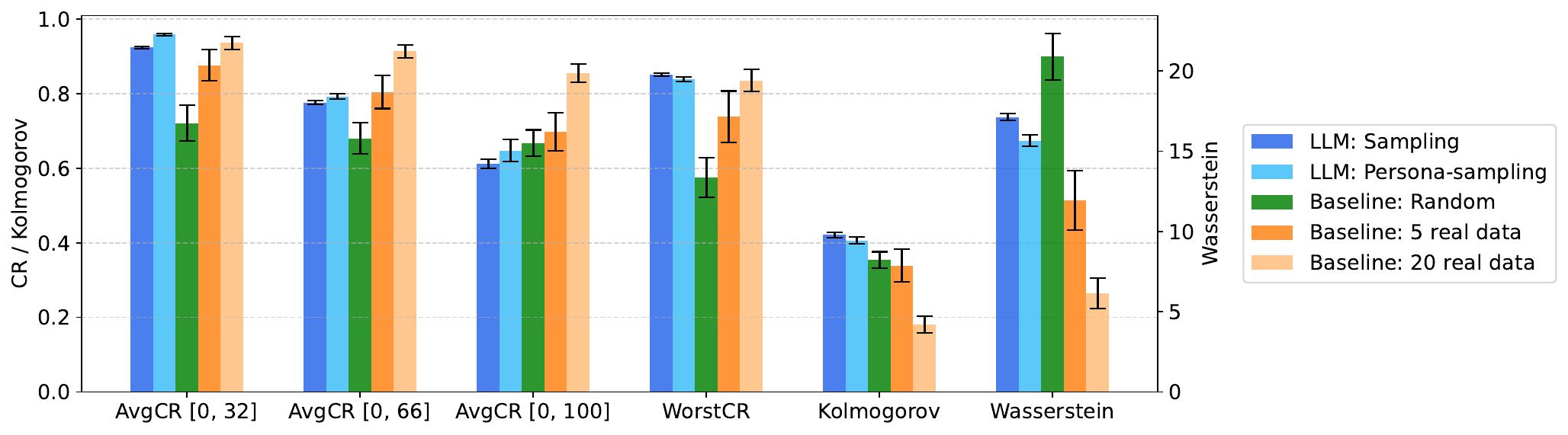}
    \caption{
    Pricing results, displaying means across 20 generations and 6 ground truths, with 95\% confidence intervals around the mean.  \textbf{Higher is better} for the CR metrics while \textbf{lower is better} for the Kolmogorov and Wasserstein metrics.  The LLM is GPT-4o.
    }
    \label{fig:pricing_bar}
\end{figure*}

\paragraph{Dataset.} We use data of willingness-to-pay for cocoa liquor in the Philippines, from \citet{ballesteros2023does}.
To elaborate, respondents bid premiums that they were willing to pay for 6 different\footnote{Technically it's 3 different products with 2 different labeling conditions (award vs.\ origin labels).} ``upgraded'' cocoa liquor products in truthful second-price auctions.
The basic cocoa liquor product had a known market price of PhP 44.00, and customer willingness-to-pay for the upgraded products is interpreted to equal 44 plus the premium they bid.
For each of the 6 products, we construct a ground-truth $F$ by taking the uniform distribution over the first 100 respondents' willingness-to-pay.

\paragraph{Details of LLM generation methods and baselines.}
As background information, we always tell the LLM the basic product description and its price of PhP 44.00, the award or origin label of the upgraded product, and importantly, the range of premiums bid (0 to 100).
In the \textbf{Few-shot} setting, we additionally give the LLM 6 randomly-drawn willingness-to-pay of respondents outside $F$.
Orthogonally, we consider all 4 ways of prompting described in \Cref{sec:generation_methods}, noting that \textbf{Persona-sampling} gives the LLM detailed characteristics of a respondent outside $F$;
\textbf{Batch-generation} uses a batch size of 25; and
\textbf{Description} asks the LLM for 5 plausible willingness-to-pay values and how it would split the probability mass among them.
Exact prompts are provided in \Cref{app:promptPricing}.
Here, the \textbf{Random} baseline is the uniform distribution over [44,144].

\paragraph{Theoretical result: computing $\worstCR(F,\hF)$.}
The problem can be formulated as minimizing $R_c(\ha)/R_c(a^*)$ over decision variables $c,a^*,\ha\ge0$, subject to~\eqref{eqn:optimalityConds}.
We show how to solve this efficiently and precisely assuming $F$ and $\hF$ have finite non-negative supports denoted by $\supp(F)$ and $\supp(\hF)$, respectively.

\begin{lemma} [proven in \Cref{pf:finiteSupp}]
\label{lem:finiteSupp}
For any cost $c$, all maximizers of $R_c(a)=(a-c)\Pr_{\xi\sim F}[\xi\ge a]$ lie within $\supp(F)$ or have $a=\infty$ (interpreted as setting a price $a>\max\{\supp(F)\}$ so that $\Pr_{\xi\sim F}[\xi\ge a]\Longrightarrow R_c(a)=0$). Analogously, $\argmax_{a\ge0}\hR_c(a)\subseteq\supp(\hF)\cup\{\infty\}$.
\end{lemma}

Therefore, constraint~\eqref{eqn:optimalityConds} is equivalent to
\begin{align*}
a^* &\in\argmax_{a\in\supp(F)\cup\{\infty\}} R_c(a);
\\ \ha &\in\argmax_{a\in\supp(\hF)\cup\{\infty\}}\hR_c(a).
\end{align*}
Now, because $R_c(a)$ is an affine function in $c$ for each fixed price $a$, we have that $\max_{a\in\supp(F)\cup\{\infty\}}R_c(a)$ is piecewise-linear in $c$, defined by breakpoints $0=:c_1<\cdots<c_n$, and that each $a\in\supp(F)\cup\{\infty\}$ lies in the $\argmax$ over an interval defined by $[c_i,c_{i'}]$ for $i,i'\in[n]$ (that can be empty if $i'<i$, or a single point if $i'=i$).
We can make analogous statements about $\hR_c(a)$ having breakpoints $0=:\hc_1<\cdots<\hc_\hn$.

\begin{lemma} [proven in \Cref{pf:mon_c}]
\label{lem:mon_c}
Given fixed values of $a^*$ and $\ha$, the function $R_c(\ha)/R_c(a^*)$ is monotonic in $c$.
\end{lemma}

Therefore, given decision variables $a^*,\ha,c$ satisfying~\eqref{eqn:optimalityConds}, if $c$ does not lie in $\{c_i\}_{i\in[n]}\cup\{\hc_i\}_{i\in[\hn]}$,
then we can improve the objective $R_c(\ha)/R_c(a^*)$ by shifting $c$ in either direction (by \Cref{lem:mon_c}), while continuing to satisfy~\eqref{eqn:optimalityConds} (by the definition of breakpoints).
For each value of $c\in\{c_i\}_{i\in[n]}\cup\{\hc_i\}_{i\in[\hn]}$, we can directly evaluate $R_c(a^*)$, while to evaluate $R_c(\ha)$ we must find the $\ha\in\argmax_{a\in\supp(\hF)\cup\{\infty\}}\hR_c(a)$ that minimizes $R_c(\ha)$.
Full details of this simple search algorithm for computing $\worstCR(F,\hF)$ are in 
\Cref{app:worstcrPricing}.

\subsection{Pricing problem: Empirical Results}

We repeat the setup from assortment (see \Cref{sec:asstEmpRes}), with full generation details and results deferred to \Cref{app:pricingTables}.
Here there are 6 products with different distributions $F$, so the reported values also take an average over these.

\Cref{fig:pricing_bar} summarizes the findings. 
LLM-based generation methods outperform the Random baseline on the decision-aware CR metrics, but less so on the decision-agnostic $\Kolmogorov$ and $\Wasserstein$ metrics.
We cap the cost parameter $c$ at 32 for $\worstCR$, and consider different ranges of $c$ for the $\avgCR$ metrics.
Higher costs create degenerate instances where the price must be set very precisely to earn any reward, making LLM-based methods less reliable at outperforming the Random baseline and unable to beat a small amount of real data (see \Cref{tab:pricing_results_all}).
Interestingly, \Cref{app:wtpPlots} reveals that the main gain of the LLM comes from generating willingness-to-pay numbers that end in 5 or 0, closely resembling those of humans (when asked for a value).

\paragraph{Is the LLM able to imitate specific personas?}\label{sec:pricingPersona}
We compute the $\mathsf{PersonaMAE}$ and $\mathsf{ShuffledMAE}$ metrics (see \Cref{sec:asstPersona}) to assess how well the LLM tailors its outputs to individual-level features.
Averaged across the six products, $\mathsf{PersonaMAE}$ and $\mathsf{ShuffledMAE}$ are \textbf{22.44} and \textbf{23.11} respectively, again reinforcing that the LLM can improve decisions despite not imitating individuals.
(For comparison, the $\mathsf{ShuffledMAE}$ of the Random baseline is \textbf{36.01}.)

\section{Newsvendor problem: Details and Results} \label{sec:newsvendor}

\paragraph{Dataset.}
We use sales data from a fashion retailer H\&M \citep{Kaggle_HM_Recommendations_2022}.
We pre-process the data to remove items with sudden discounts or inventory stockouts (details in \Cref{app:invPreprocessing}), so that the aggregate purchases reflect total demands under a stable price.
We take all products with type \textit{trouser},  which is the largest type with 300 items after the pre-processing. 
For each item, we take their weekly demand over 30 weeks, and use this empirical distribution to form $F$. 
Each item is associated with metadata such as the product name, type, and color (details in \Cref{app:invPreprocessing}).

\paragraph{Details of LLM generation methods and baselines.}
We utilize the \textbf{Descriptive} generation method with Few-shot examples of other distributions.
Specifically, for a target item, the LLM is asked predict the mean and standard deviation for a normal distribution corresponding to the item's weekly demand.
The LLM is provided with  the target item's metadata, and also a reference set of 100 other example items, including the metadata and historical demand parameters (mean and standard deviation) for each reference item.

We do not employ the other generation methods because this application is structurally different than the previous settings. In assortment and pricing, each data point represented the choice or valuation of a \textit{single individual}, hence those settings were aligned with the literature on using LLMs to simulate humans, which motivates the sampling-based methods.
In newsvendor, a single observation is a weekly demand realization for an item, which makes the ``human simulation'' paradigm less applicable, and hence we omit the agent-based sampling methods.

We compare against a \textbf{Random} baseline which uses the empirical distribution of the demands of \textit{all} 300 items.

\paragraph{Theoretical result: computing $\worstCR(F,\hF)$.}
The problem can be formulated as minimizing $R_q(a^*)/R_q(\ha)$ over decision variables $q\in(0,1)$ and $a^*,\ha\ge0$, subject to~\eqref{eqn:optimalityConds}.
We show how to solve this efficiently and precisely assuming $F$ and $\hF$ are discrete uniform distributions over $m$ and $\hm$ real numbers, respectively.

\begin{lemma}[proven in \Cref{pf:shift_q}] \label{lem:shift_q}
If $a^*\in\argmax_{a\ge0} R_q(a)$ for some $q\in(\frac{i-1}m,\frac im)$ and $i\in[m]$, then $a^*$ continues to lie in the $\argmax$ for all $q\in[\frac{i-1}m,\frac im]$.
Analogously, if $\ha\in\argmax_{a\ge0} \hR_q(a)$ for some $q\in(\frac{i-1}{\hm},\frac i{\hm})$ and $i\in[\hm]$, then $\ha$ lies in the $\argmax$ for all $q\in[\frac{i-1}{\hm},\frac i{\hm}]$.
\end{lemma}

\begin{lemma}[proven in \Cref{pf:mon_q}] \label{lem:mon_q}
Given fixed values of $a^*$ and $\ha$, the function $R_q(a^*)/R_q(\ha)$ is monotonic in $q$.
\end{lemma}

Given decision variables $a^*,\ha,q$ satisfying~\eqref{eqn:optimalityConds}, if $q$ is a multiple of neither $1/m$ nor $1/\hm$, then we can improve the objective $R_q(a^*)/R_q(\ha)$ by shifting $q$ in either direction (by \Cref{lem:mon_q}), while continuing to satisfy~\eqref{eqn:optimalityConds} (by \Cref{lem:shift_q}).
Then, we can restrict the search to $q\in\{\frac1m,\ldots,\frac{m-1}m\}\cup\{\frac1{\hm},\ldots,\frac{\hm-1}{\hm}\}$.
For each such value of $q$, we must find the $\ha\in\argmax_{a\ge0} \hR_q(a)$ that minimizes $R_q(\ha)$.
It can be shown that $\argmax_{a\ge0} \hR_q(a)$ is an interval and $R_q(\cdot)$ is unimodal, and hence it suffices to check the endpoints of the interval.  Full algorithm details are in \Cref{app:worstcrNews}.

\subsection{Newsvendor problem: Empirical Results}

\begin{figure*}[h]
\centering
        \includegraphics[width=0.7\textwidth]{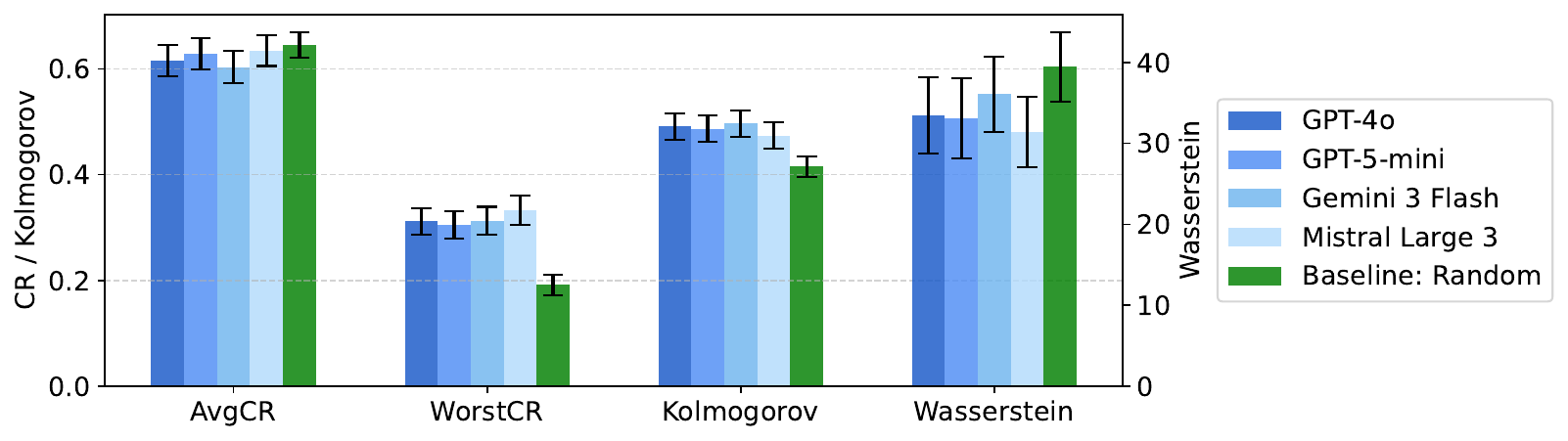}
    \caption{
    Newsvendor results across four LLM models, displaying means across 300 items with 95\% confidence intervals.  \textbf{Higher is better} for the CR metrics while \textbf{lower is better} for the Kolmogorov and Wasserstein metrics.  
    The
    $\avgCR$ uses the distribution $q \sim \mathrm{Unif}[0.01, 0.99]$.
    }
    \label{fig:newsvendor_bar}
\end{figure*}

\Cref{fig:newsvendor_bar} reports results for the CR metrics and decision-agnostic distributional distances for four LLM models, averaging across the 300 items.
We see that all LLM models perform similarly to the Random baseline for $\avgCR$, as all confidence bars are overlapping, but LLMs improve upon $\worstCR$ substantially.
The two decision-unaware distances point in opposite directions (Random baseline is better under Kolmogorov, but worse under Wasserstein), suggesting that  decision-unaware metrics can be unreliable indicators of downstream decision quality.

\section{Concluding Discussion}



Our work flows into an emerging stream on using LLMs to make business-relevant decisions, and makes progress toward the important goal of benchmarking LLM progress at this task. 
More specifically, we evaluate how well LLMs can use their world knowledge to generate data/distributions that feed into an SAA optimizer, and we find that the performance is promising.

At the same time, biased, unrepresentative, or over-trusted LLM-generated distributions could lead to poor or unfair decisions that impact customers, so deployment would require additional domain-specific validation and fairness auditing.

\bibliographystyle{abbrvnat}
\bibliography{bibliography}

\newpage
\appendix

\section{Deferred Proofs}

\subsection{Proof of \Cref{lem:asstKey}} \label{pf:asstKey}

First suppose $\min\{|a^*\setminus\ha|,|\ha\setminus a^*|\}\le 1$.  Without loss, let $|a^*\setminus\ha|\le1$.  Set $B=1$. For all $j\in[n]$, set
\begin{align*}
c_j=
\begin{cases}
0, &\text{ if } j\in a^*\cap\ha;
\\ 2, &\text{ if } j\notin a^*\cup\ha;
\\ 1, &\text{ if } j\in a^*\setminus\ha;
\\ 1/n, &\text{ if } j\in\ha\setminus a^*.
\end{cases}
\end{align*}
Under the resulting $\cA_\theta$, both $a^*$ and $\ha$ is feasible (because $|a^*\setminus\ha|\le1$).
Moreover, elements outside $a^*\cup\ha$ can never be included.
Thus, for a set $a\notin\{a^*,\ha\}$ to be maximal, it needs to contain elements from both $a^*\setminus\ha$ and $\ha\setminus a^*$.
But any such set would have total size exceeding 1, completing the proof that only the sets $a^*,\ha$ can be maximal in $\cA_\theta$.

Conversely, suppose $\min\{|a^*\setminus\ha|,|\ha\setminus a^*|\}\ge2$.
For distinct items $j_1,j_2\in a^*\setminus\ha$ with $s_{j_1}\le s_{j_2}$, we know $s_{j_1}+s_{j_2}+\sum_{j\in a^*\cap\ha} s_j\le B$, and hence $2s_{j_1}+\sum_{j\in a^*\cap\ha} s_j\le B$.
Similarly, for distinct items $j_3,j_4\in \ha\setminus a^*$ with $s_{j_3}\le s_{j_4}$, we know $2s_{j_3}+\sum_{j\in a^*\cap\ha} s_j\le B$.
This collectively implies $s_{j_1}+s_{j_3}+\sum_{j\in a^*\cap\ha} s_j\le B$, and hence $\{j_1,j_3\}\cup(a^*\cap\ha)$ (or a set containing it) must be maximal in $\cA_\theta$.

\subsection{Proof of \Cref{lem:finiteSupp}} \label{pf:finiteSupp}

Let $F$ be a discrete distribution supported on
$\mathrm{supp}(F)=\{\xi^1,\ldots,\xi^m\}$,
where $0 \le \xi^1 < \cdots < \xi^m$, and $\xi^0$ is understood to be 0. Without loss of generality, restrict $a>0$.
First suppose $a \in (\xi^n,\xi^{n+1}]$ for some $n\in\{0,\ldots,m-1\}$. We know $\Pr_{\xi\sim F}[\xi\ge a]$ is constant on this interval, hence $\argmax_{a\in(\xi^n,\xi^{n+1}]} R_c(a)=\xi^{n+1}$. Next, when $a>\max\{\supp(F)\}$, $\Pr_{\xi\sim F}[\xi\ge a]=0$, then $a$ can be set to $\infty$. Thus, $\argmax_{a\ge0}R_c(a)\subseteq\supp(F)\cup\{\infty\}$. The result for $\argmax_{a\ge0} \hR_c(a)$ is proved analogously based on $\hF$.

\subsection{Proof of \Cref{lem:mon_c}} \label{pf:mon_c}
Fix $a^*, \hat a \ge 0$ and write $D^* := \Pr_{\xi\sim F}[\xi \ge a^*]$, $\hat D := \Pr_{\xi\sim F}[\xi \ge \hat a]$. Then 
\[
\frac{R_c(\hat a)}{R_c(a^*)}
=
\frac{(\hat a - c)\hat D}{(a^* - c)D^*}
=
\frac{\alpha + \beta c}{\gamma + \delta c},
\]
where $\alpha := \hat a \hat D$, $\beta := -\hat D$, $\gamma := a^* D^*$, $\delta := -D^*$.
On any interval where $\gamma + \delta c \neq 0$, the derivative satisfies
\[
\frac{d}{dc}\!\left(\frac{\alpha + \beta c}{\gamma + \delta c}\right)
=
\frac{\beta\gamma - \alpha\delta}{(\gamma + \delta c)^2},
\]
whose sign is constant. Hence, $R_c(\hat a)/R_c(a^*)$ is monotone in $c$
(or constant in the degenerate case).

\subsection{Proof of \Cref{lem:shift_q}} \label{pf:shift_q}

Let $F$ be uniform over values $\xi^1,\ldots,\xi^m$ where $0\le\xi^1\le\cdots\le\xi^m$.
It can be checked that
\begin{align*}
\argmax_{a\ge0} R_q(a)
&=
\begin{cases}
\xi^i, &\text{ if } q\in(\frac{i-1}m,\frac im),i\in[m];
\\ [\xi^i,\xi^{i+1}], &\text{ if } q=\frac im,i\in[m]\cup\{0\};
\end{cases}
\end{align*}
where $\xi^0$ is understood to be 0 and $\xi^{m+1}$ is understood to be $\infty$.
From this the result about $\argmax_{a\ge0} R_q(a)$ follows.
The result for $\argmax_{a\ge0} \hR_q(a)$ is proved analogously based on $\hF$.

\subsection{Proof of \Cref{lem:mon_q}} \label{pf:mon_q}

Fix $a^*,\hat a\ge 0$ and define $U^*:=\mathbb{E}[[\xi-a^*]^+]$, $O^*:=\mathbb{E}[[a^*-\xi]^+]$,
$\hat U:=\mathbb{E}[[\xi-\hat a]^+]$, and $\hat O:=\mathbb{E}[[\hat a-\xi]^+]$ to be the expected understockings and overstockings of $a^*$ and $\ha$ respectively.
Then
\[
\frac{R_q(a^*)}{R_q(\hat a)}
=\frac{qU^*+(1-q)O^*}{q\hat U+(1-q)\hat O}
=\frac{O^*+q(U^*-O^*)}{\hat O+q(\hat U-\hat O)}
=\frac{\alpha+\beta q}{\gamma+\delta q},
\]
where $\alpha=O^*$, $\beta=U^*-O^*$, $\gamma=\hat O$, $\delta=\hat U-\hat O$.
On any interval where $\gamma+\delta q\neq 0$, the derivative is
\[
\frac{d}{dq}\!\left(\frac{\alpha+\beta q}{\gamma+\delta q}\right)
=\frac{\beta\gamma-\alpha\delta}{(\gamma+\delta q)^2},
\]
whose sign is constant; hence the ratio is monotone in $q$ (or constant in the degenerate case).

\clearpage
\section{Computing $\worstCR(F,\hF)$ for Assortment} \label{app:worstcrAsst}

\begin{figure}[H]
\centering
\includegraphics[width=0.6\linewidth]{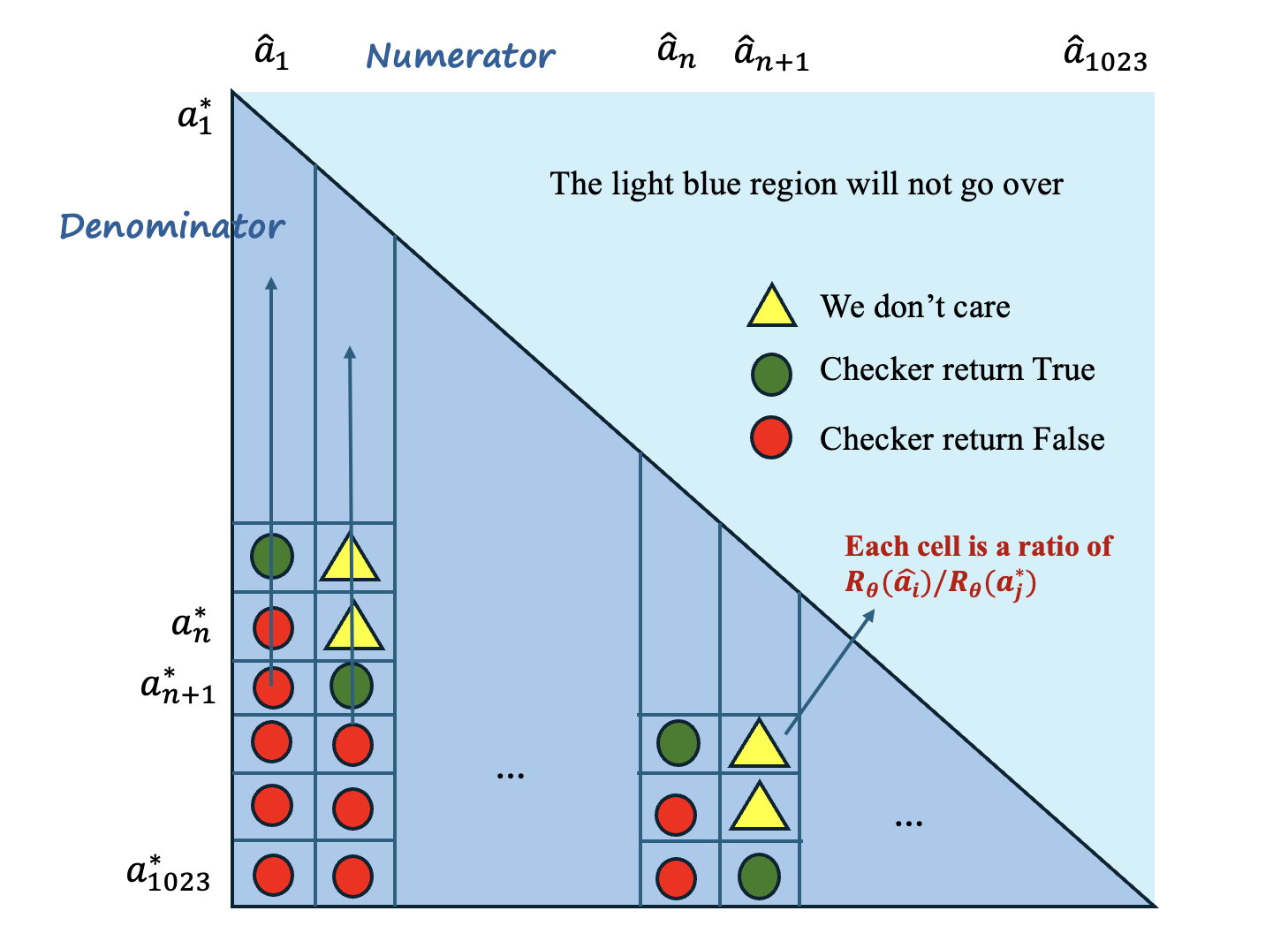}
\caption{Illustration of the enumeration and pruning procedure, in our special case where $n=10$ and hence there are 1023 non-empty assortments. Rows correspond to $R_\theta(\hat a)$ and columns correspond to $R_\theta(a^\ast)$
, while each cell represents a ratio $R_\theta(\hat a)/R_\theta(a^\ast)$. Assortments are ordered by $R_\theta(\cdot)$, with $\hat a_1$ and $a^\ast_1$ denoting the smallest rewards in their respective orders. We traverse only the red and green cells; light-blue cells violate \eqref{eqn:optimalityConds} and are pruned. The boundary is illustrative and not necessarily diagonal.}
\label{fig:assortment_pruning} 
\end{figure}

To find $\worstCR(F,\hat F)$, we enumerate candidate ratios only over pairs
$(a^\ast,\hat a)$ that satisfy \eqref{eqn:theoryPrereq} to check whether \eqref{eqn:optimalityConds} holds, which allows us to discard nearly half of all candidate pairs. Furthermore, here is a pruning trick for not checking all the valid pairs of \eqref{eqn:theoryPrereq}: we perform a double-loop search, with $R_\theta(\hat a)$ ordered increasingly and $R_\theta(a^\ast)$ ordered decreasingly. For a fixed numerator $R_\theta(\hat a)$, we iterate over candidate denominators $R_\theta(a^\ast)$ and check whether there exists a parameter $\theta$ such that \eqref{eqn:optimalityConds} holds. Once such a $\theta$ is found, we terminate the inner loop for this $\hat a$, since under the current enumeration order any subsequent denominator would yield a larger ratio
and cannot improve the result.
If the corresponding ratio $R_\theta(\hat a)/R_\theta(a^\ast)$ is strictly smaller than the current $\worstCR$, we update it accordingly, breaking ties by choosing the pair with the smaller $R_\theta(\hat a)$. Figure\ref{fig:assortment_pruning} illustrates this pruning procedure.

For validity checking, if $\min\{|a^\ast \setminus \hat a|,\;|\hat a \setminus a^\ast|\} \le 1$, then by Lemma\ref{lem:asstKey}, \eqref{eqn:optimalityConds} holds. If $\min\{|a^\ast \setminus \hat a|,\;|\hat a \setminus a^\ast|\} > 1$, a candidate ratio can be invalid
if there exists an assortment $\hat{a}_{\text{bad}}$ that satisfies $R_\theta(a_{\text{bad}}) > R_\theta(a^\ast)  \,
\text{or}\,
 \hat R_\theta(a_{\text{bad}}) > \hat R_\theta(\hat a)$ and there do not exist sizes $(s_j)_{j\in[n]}$ and a budget $B$ that can exclude all ``bad'' assortments from $\cA_\theta$.  In this case,\eqref{eqn:optimalityConds} can't hold and the corresponding ratio $R_\theta(\hat a)/R_\theta(a^\ast)$ cannot actually be realized by the adversary. 

 To determine whether such sizes $(s_j)_{j\in[n]}$ and a budget $B$ that can exclude all ``bad'' assortments exist, we formulate a linear feasibility program:

\[
\begin{aligned}
\sum_{i=1}^n s_i\, x_i(a^\ast) &\le B, \\
\sum_{i=1}^n s_i\, x_i(\hat a) &\le B, \\
\sum_{i=1}^n s_i\, x_i(a_{\text{bad}}) &> B,
\quad \forall\, a_{\text{bad}} \\
s_i &\ge 0, \quad i=1,\dots,n .
\end{aligned}
\]

Here $x(a)\in\{0,1\}^n$ is the indicator vector of assortment $a$
($x_i(a)=1$ if item $i$ is included, and $0$ otherwise).

The overall algorithm is given in Algorithm\ref{alg:worst-cr}.

\begin{algorithm}[H]
\caption{\textsc{Worst-Case Competitive Ratio Search for Assortment}}
\label{alg:worst-cr}
\begin{algorithmic}[1]
\REQUIRE Ground-truth distribution $F$ and estimation distribution $\hF$
\ENSURE WorstCR correspond to each $\hF$
\STATE \textbf{Step 1: Compute rewards and sort subsets}
\FOR{$F$, $\hF$} 
    \FOR{each subset $a_{\text{temp}}\subseteq[n]$}
        \STATE Compute reward $R(a_{\text{temp}}^\ast)$, $R(\ha_{\text{temp}})$;
    \ENDFOR
\ENDFOR
\STATE Sort all subsets in ascending order of $R(a_{\text{temp}}^\ast)$ and $R(\ha_{\text{temp}})$ to obtain list $\mathcal{L^*} = \{ a_1^*, a_2^*, \ldots, a_{1023}^* \}$ and $\mathcal{L}^{\ha} = \{ \ha_1, \ha_2, \ldots, \ha_{1023} \}$.
\STATE \textbf{Step 2: Find WorstCR}
\FOR{{each $\hF$}}
    \STATE Initialize $\mathrm{lower\_bound} \gets 1$.
        \FOR{$\ha_i$ in $\mathcal{L}^{\ha}$ from smallest reward to largest}
            \FOR{$a_j^*$ in $\mathcal{L^*}$ from largest reward to smallest}
                \STATE Compute ratio $\rho = R(\ha_i)/R(a^\ast_j)$;
                    \IF{$\rho \ge \mathrm{lower\_bound}$}
                    \STATE continue.
                    \ENDIF
                    \IF{ $\hR_\theta(\ha_i)<\hR_\theta(a^*_j)$}
                    \STATE continue.
                    \ENDIF
                    \IF{$\min\{|a^\ast_j \setminus \ha_i|,\;|\ha_i \setminus a^\ast_j|\} \le 1$}
                    \STATE Update $\mathrm{lower\_bound} \gets \rho$.
                    \STATE break.
                    \ELSE 
                    \STATE Initialize set of $a_\text{bad} \gets \emptyset$.
                    \FOR{$a_{\text{temp}} \subseteq \mathcal{P}(a_j^* \cup \ha_i)$}
                        \IF{$R(a_{\text{temp}}) > R(a^\ast)  \,\text{or}\,\hR(a_{\text{temp}}) > \hR(\ha)$}
                        \STATE Add $a_{\text{temp}}$ to the set of $a_\text{bad}$
                        \ENDIF
                    \ENDFOR
                    \FOR{$a_{\text{bad}}$ in the set of $a_{\text{bad}}$}
                        \IF{LP is feasible}
                            \STATE Update $\mathrm{lower\_bound} \gets \rho$.
                            \STATE break.
                        \ENDIF
                    \ENDFOR
                    \ENDIF
            \ENDFOR
        \ENDFOR
\ENDFOR

\STATE\RETURN $\mathrm{lower\_bound}$

\end{algorithmic}
\end{algorithm}

\section{Computing $\worstCR(F,\hF)$ for Pricing} \label{app:worstcrPricing}


\begin{algorithm}[H]
\caption{\textsc{Worst-Case Competitive Ratio Search for Pricing}}
\label{alg:worstcr-pricing}
\begin{algorithmic}[1]
\scriptsize
\REQUIRE Ground-truth distribution $F$ and estimation distribution $\hF$.
\ENSURE $\mathrm{WorstCR}(F,\hF)$
\STATE $x_{\min}\gets 0$
\STATE \textbf{Step1: Build affine-line families}
\STATE \hspace{1em} $\mathcal{L}_F=\{\ell_a(c)=(a-c)\Pr_{\xi\sim F}[\xi\ge a] : a\in \mathrm{supp}(F)\cup\{\infty\}\}$. 
\STATE \hspace{1em} $\mathcal{L}_{\hF}=\{\widehat \ell_a(c)=(a-c)\Pr_{\xi\sim \hF}[\xi\ge a] : a\in \mathrm{supp}(\hF)\cup\{\infty\}\}$. 
\STATE $\mathcal{C} \gets \{0\}$
\STATE \textbf{Step2: Find Upper Envelope}
\FOR{each family $\mathcal{L} \in \{\mathcal{L}_F, \mathcal{L}_{\hF}\}$}
\STATE $\texttt{Stack} \gets \emptyset$; $\texttt{BreakPtr} \gets \emptyset$.
    \STATE Sort $\mathcal{L}$ by slope (tail prob.) descending; handle ties by keeping max intercept.

    \FOR{each line $\ell_j \in \mathcal{L}$}

    \IF{$\texttt{Stack}$ is empty}
        \STATE push $\ell_j$ to $\texttt{Stack}$
        \STATE \textbf{continue} 
    \ENDIF
    \STATE $x^\star \gets \text{intersect}(\ell_j, \text{top}(\texttt{Stack}))$
    \IF{$|\texttt{Stack}|=1$ \AND $x^\star \le x_{\min}$}
    \STATE pop \texttt{Stack} 
    \STATE push $\ell_j$ to $\texttt{Stack}$
    \STATE \textbf{continue}
    \ENDIF
    \WHILE{$|\texttt{Stack}| \ge 2$}
      \STATE $x^\star \gets \text{intersect}(\ell_j, \text{top}(\texttt{Stack}))$ %
      \IF{$x^\star \le \text{last}(\texttt{BreakPtr})$}
        \STATE pop \texttt{Stack}; pop \texttt{BreakPtr}
      \ELSE
        \STATE \textbf{break}
      \ENDIF
    \ENDWHILE

    \STATE $x^\star \gets \text{intersect}(\ell_j, \text{top}(\texttt{Stack}))$ 
    \STATE push $x^\star$ to \texttt{BreakPtr}; push $\ell_j$ to \texttt{Stack}
\ENDFOR
\STATE $\mathcal{C} \gets \mathcal{C} \cup \{x \in \texttt{BreakPtr} \mid x \ge x_{\min}\}$. 

  \IF{$\mathcal{L}=\mathcal{L}_F$}
    \STATE $\texttt{Stack}_F \gets \texttt{Stack}$; \ $\texttt{BreakPtr}_F \gets \texttt{BreakPtr}$ 
  \ELSE 
    \STATE $\texttt{Stack}_{\hF} \gets \texttt{Stack}$; \ $\texttt{BreakPtr}_{\hF} \gets \texttt{BreakPtr}$
  \ENDIF 
\ENDFOR

\STATE \textbf{Step 3: WorstCR Search}
\STATE Sort $\mathcal{C}$ and remove duplicates: $c_1 < c_2 < \dots < c_m$.
\STATE $p \gets 1, \hat{p} \gets 1$; $\text{WorstRatio} \gets \infty$.
\FOR{each $c_t \in \mathcal{C}$}

    \WHILE{$p \le |\texttt{BreakPtr}_F|$ \AND $\texttt{BreakPtr}_F[p] \le c_t$} 
    \STATE $p \gets p + 1$ \ENDWHILE
    
    \WHILE{$\hat{p} \le |\texttt{BreakPtr}_{\hF}|$ \AND $\texttt{BreakPtr}_{\hF}[\hat{p}] \le c_t$} 
    \STATE $\hat{p} \gets \hat{p} + 1$ 
  \ENDWHILE
    
    \STATE $a^\star \gets \text{price of } \texttt{Stack}_F[p]$
    \STATE $\widehat{A} \gets \{ \text{price of } \texttt{Stack}_{\hF}[\hat{p}] \}$
    
    \IF{$\hat{p} \le |\texttt{BreakPtr}_{\hF}|$ \AND $c_t = \texttt{BreakPtr}_{\hF}[\hat{p}]$} 
    \STATE $\widehat{A} \gets \widehat{A} \cup \{\,\text{price of } \texttt{Stack}_{\hF}[\hat{p}+1]\,\}$ 
  \ENDIF
    
    \STATE $\text{WorstRatio} \gets \min \left( \text{WorstRatio}, \frac{\min_{a \in \widehat{A}} (a-c_t)\Pr_{\xi\sim F}[\xi\ge a]}{(a^\star-c_t)\Pr_{\xi\sim F}[\xi\ge a^\star]} \right)$
\ENDFOR
\RETURN $\text{WorstRatio}$.

\end{algorithmic}
\end{algorithm}

\clearpage
\section{Computing $\worstCR(F,\hF)$ for Newsvendor} \label{app:worstcrNews}

\begin{algorithm}[H]
\caption{\textsc{WorstCR-Newsvendor}$(F,\hF)$ (discrete uniform supports)}
\label{alg:worstcr-newsvendor}
\begin{algorithmic}[1]
\REQUIRE $F$ uniform over $m$ demand values; $\hF$ uniform over $\hm$ demand values.
\ENSURE $\mathrm{WorstCR}(F,\hF)$ over $q\in(0,1)$.

\STATE Define rewards (negative loss):
\STATE \hspace{1em} $R_q(a)= -\left(q\,\mathbb{E}_{\xi\sim F}[(\xi-a)^+] + (1-q)\,\mathbb{E}_{\xi\sim F}[(a-\xi)^+]\right)$,
\STATE \hspace{1em} and analogously $\hR_q(a)$ under $\hF$.
\STATE Candidate parameters:
\STATE \hspace{1em} $\mathcal{Q}\leftarrow \{1/m,\ldots,(m-1)/m\}\ \cup\ \{1/\hm,\ldots,(\hm-1)/\hm\}$.

\STATE $\texttt{best}\leftarrow +\infty$.
\FOR{each $q\in\mathcal{Q}$}
  \STATE Compute $a^\star \in \arg\max_{a\ge 0} R_q(a)$.
  \STATE Compute $\widehat A(q)=\arg\max_{a\ge 0}\widehat R_q(a)$.
  \STATE Let $\underline a=\min \widehat A(q)$ and $\overline a=\max \widehat A(q)$.
  \STATE $\widehat a \leftarrow \arg\min\{R_q(\underline a),\,R_q(\overline a)\}$.
  \STATE $\texttt{ratio}\leftarrow \dfrac{R_q(a^\star)}{R_q(\widehat a)}$.
  \STATE $\texttt{best}\leftarrow \min(\texttt{best},\texttt{ratio})$.
\ENDFOR
\STATE \textbf{return} $\texttt{best}$.
\end{algorithmic}
\end{algorithm}

\clearpage
\section{Assortment problem: Generation Details and Full Results}
\label{app:asstTables}

We consider 20 different distributions $\hF$ for each LLM generation method and each baseline, calling them multiple times to handle the randomness in the distribution generated.
\begin{itemize}
\item \textbf{Sampling}: We generate a pool of 600 rankings using the LLM. We subsample 200 rankings to form an estimated distribution $\hF$, and repeat to form 20 estimated distributions in this way.
\item \textbf{Persona-sampling}:  We generate a pool of 600 rankings, prompting the LLM with a different persona each time. We subsample 200 rankings to form an estimated distribution $\hF$, and repeat to form 20 estimated distributions in this way.
\item \textbf{Batch-generation}: We generate 30 rankings per query and repeat this process 20 times, yielding 20 distributions. For each probability distribution, we sample 50 rankings according to the specified proportions, resulting in 20 distributions.
\item \textbf{Description}: We ask the LLM to generate a score vector for the sushi items, repeating 20 times. For each score vector, we use a Plackett–Luce model to sample 200 rankings, resulting in 20 distributions.
\item \textbf{Few-shot Sampling}: We draw 5 example sets, each containing 6 examples. For each example set, we generate a pool of 300 rankings and construct 4 subsamples of size 200, resulting in a total of 20 distributions.
\item \textbf{Few-shot Persona-sampling}: Same as Few-shot Sampling, except we also prompt the LLM with a different persona each time.
\item \textbf{Few-shot Batch-generation}: We draw 5 example sets, each containing 6 examples.  We run Batch-generation 4 times with each example set, resulting in a total of 20 distributions.
\item \textbf{Few-shot Description}: We draw 5 example sets, each containing 6 examples. We run Description 4 times with each example set, resulting in a total of 20 distributions.
\item \textbf{Random baseline}: We generate a pool of 600 rankings uniformly at random.  We subsample 200 rankings from it 20 times, to form 20 estimated distributions.
\item \textbf{$d$ real data baseline}: We randomly sample $d$ rankings from the ground-truth $F$, generating 20 distributions in this way.
\end{itemize}

This appendix reports three versions of the assortment results table, which differ only in the choice of the position-dependent rewards $(r_p)_{p\in[n]}$ used in the reward function $r_\theta(a,\xi)=r_{\min\{p\in[n]:\xi_p\in a\}}$.
Specifically, the three tables use $r_p=10/p$, $r_p=10-(p-1)$, and $r_p=10-0.1(p-1)^2$, respectively.

\begin{table*}[!htbp]
\footnotesize
\centering
\begin{tabular}{l l c c c c c c c c}
\toprule
 && \multirow{1}{*}{\textbf{Wasserstein}} & \multirow{1}{*}{\textbf{WorstCR}} & \multicolumn{6}{c}{\textbf{AverageCR} \textit{(higher is better)}} \\
 \cmidrule(lr){5-10}
 && \textit{(lower is}  & \textit{(higher  is} & \multicolumn{2}{c}{Unit} & \multicolumn{2}{c}{Random} & \multicolumn{2}{c}{Hard} \\
\textbf{Model} & \textbf{Method} & \textit{better)} &  \textit{better)} & $B=2$  & $B=5$  & $B=2$  & $B=5$  & $B=2$  & $B=5$ \\
\midrule \multirow{8}{*}{GPT-4o} & Sampling & \cellcolor[RGB]{47,127,188}\textcolor{white}{0.23} & \cellcolor[RGB]{169,207,229}\textcolor{black}{0.42} & \cellcolor[RGB]{8,62,129}\textcolor{white}{0.96} & \cellcolor[RGB]{51,131,190}\textcolor{white}{0.93} & \cellcolor[RGB]{36,116,183}\textcolor{white}{0.91} & \cellcolor[RGB]{81,156,204}\textcolor{black}{0.92} & \cellcolor[RGB]{143,194,222}\textcolor{black}{0.74} & \cellcolor[RGB]{8,74,145}\textcolor{white}{0.95} \\
 & Sampling, Few-shot & \cellcolor[RGB]{42,122,185}\textcolor{white}{0.23} & \cellcolor[RGB]{191,216,237}\textcolor{black}{0.39} & \cellcolor[RGB]{8,57,121}\textcolor{white}{0.97} & \cellcolor[RGB]{35,115,182}\textcolor{white}{0.94} & \cellcolor[RGB]{42,122,185}\textcolor{white}{0.90} & \cellcolor[RGB]{50,130,190}\textcolor{white}{0.94} & \cellcolor[RGB]{72,150,200}\textcolor{white}{0.82} & \cellcolor[RGB]{8,74,145}\textcolor{white}{0.95} \\
 & Persona & \cellcolor[RGB]{48,128,189}\textcolor{white}{0.24} & \cellcolor[RGB]{148,196,223}\textcolor{black}{0.44} & \cellcolor[RGB]{8,62,129}\textcolor{white}{0.96} & \cellcolor[RGB]{8,52,113}\textcolor{white}{0.99} & \cellcolor[RGB]{8,65,132}\textcolor{white}{0.98} & \cellcolor[RGB]{8,69,138}\textcolor{white}{0.98} & \cellcolor[RGB]{8,48,107}\textcolor{white}{1.00} & \cellcolor[RGB]{8,67,135}\textcolor{white}{0.96} \\
 & Persona, Few-shot & \cellcolor[RGB]{28,106,176}\textcolor{white}{0.22} & \cellcolor[RGB]{154,200,224}\textcolor{black}{0.43} & \cellcolor[RGB]{8,59,124}\textcolor{white}{0.97} & \cellcolor[RGB]{8,62,129}\textcolor{white}{0.98} & \cellcolor[RGB]{36,116,183}\textcolor{white}{0.91} & \cellcolor[RGB]{8,81,156}\textcolor{white}{0.97} & \cellcolor[RGB]{72,150,200}\textcolor{white}{0.82} & \cellcolor[RGB]{17,92,165}\textcolor{white}{0.94} \\
 & Batch & \cellcolor[RGB]{234,242,251}\textcolor{black}{0.34} & \cellcolor[RGB]{208,225,242}\textcolor{black}{0.37} & \cellcolor[RGB]{109,175,215}\textcolor{black}{0.83} & \cellcolor[RGB]{114,178,216}\textcolor{black}{0.89} & \cellcolor[RGB]{163,204,227}\textcolor{black}{0.78} & \cellcolor[RGB]{96,167,210}\textcolor{black}{0.91} & \cellcolor[RGB]{166,206,228}\textcolor{black}{0.71} & \cellcolor[RGB]{42,122,185}\textcolor{white}{0.91} \\
 & Batch, Few-shot & \cellcolor[RGB]{238,245,252}\textcolor{black}{0.34} & \cellcolor[RGB]{217,231,245}\textcolor{black}{0.35} & \cellcolor[RGB]{205,224,241}\textcolor{black}{0.75} & \cellcolor[RGB]{200,220,240}\textcolor{black}{0.84} & \cellcolor[RGB]{171,208,230}\textcolor{black}{0.77} & \cellcolor[RGB]{168,206,228}\textcolor{black}{0.87} & \cellcolor[RGB]{175,209,231}\textcolor{black}{0.69} & \cellcolor[RGB]{191,216,237}\textcolor{black}{0.82} \\
 & Description & \cellcolor[RGB]{8,66,133}\textcolor{white}{0.20} & \cellcolor[RGB]{25,102,173}\textcolor{white}{0.58} & \cellcolor[RGB]{8,79,153}\textcolor{white}{0.95} & \cellcolor[RGB]{21,98,169}\textcolor{white}{0.96} & \cellcolor[RGB]{21,98,169}\textcolor{white}{0.93} & \cellcolor[RGB]{27,105,175}\textcolor{white}{0.96} & \cellcolor[RGB]{116,179,216}\textcolor{black}{0.76} & \cellcolor[RGB]{8,81,156}\textcolor{white}{0.95} \\
 & Description, Few-shot & \cellcolor[RGB]{8,61,127}\textcolor{white}{0.19} & \cellcolor[RGB]{61,141,196}\textcolor{white}{0.52} & \cellcolor[RGB]{12,86,160}\textcolor{white}{0.94} & \cellcolor[RGB]{29,108,177}\textcolor{white}{0.95} & \cellcolor[RGB]{28,106,176}\textcolor{white}{0.92} & \cellcolor[RGB]{42,122,185}\textcolor{white}{0.94} & \cellcolor[RGB]{60,140,195}\textcolor{white}{0.84} & \cellcolor[RGB]{55,135,192}\textcolor{white}{0.90} \\
\midrule \multirow{8}{*}{GPT-5-mini} & Sampling & \cellcolor[RGB]{193,217,237}\textcolor{black}{0.30} & \cellcolor[RGB]{148,196,223}\textcolor{black}{0.44} & \cellcolor[RGB]{8,62,129}\textcolor{white}{0.96} & \cellcolor[RGB]{166,206,228}\textcolor{black}{0.86} & \cellcolor[RGB]{8,73,144}\textcolor{white}{0.97} & \cellcolor[RGB]{130,187,219}\textcolor{black}{0.89} & \cellcolor[RGB]{8,48,107}\textcolor{white}{1.00} & \cellcolor[RGB]{20,96,168}\textcolor{white}{0.93} \\
 & Sampling, Few-shot & \cellcolor[RGB]{99,168,211}\textcolor{black}{0.26} & \cellcolor[RGB]{106,174,214}\textcolor{black}{0.47} & \cellcolor[RGB]{23,100,171}\textcolor{white}{0.92} & \cellcolor[RGB]{10,83,158}\textcolor{white}{0.97} & \cellcolor[RGB]{11,85,159}\textcolor{white}{0.95} & \cellcolor[RGB]{19,95,167}\textcolor{white}{0.96} & \cellcolor[RGB]{8,78,152}\textcolor{white}{0.95} & \cellcolor[RGB]{8,67,135}\textcolor{white}{0.96} \\
 & Persona & \cellcolor[RGB]{22,99,170}\textcolor{white}{0.22} & \cellcolor[RGB]{199,220,239}\textcolor{black}{0.38} & \cellcolor[RGB]{247,251,255}\textcolor{black}{0.69} & \cellcolor[RGB]{20,96,168}\textcolor{white}{0.96} & \cellcolor[RGB]{183,212,234}\textcolor{black}{0.76} & \cellcolor[RGB]{34,114,182}\textcolor{white}{0.95} & \cellcolor[RGB]{148,196,223}\textcolor{black}{0.73} & \cellcolor[RGB]{8,53,115}\textcolor{white}{0.97} \\
 & Persona, Few-shot & \cellcolor[RGB]{25,102,173}\textcolor{white}{0.22} & \cellcolor[RGB]{175,209,231}\textcolor{black}{0.41} & \cellcolor[RGB]{75,152,202}\textcolor{white}{0.86} & \cellcolor[RGB]{8,64,130}\textcolor{white}{0.98} & \cellcolor[RGB]{104,172,213}\textcolor{black}{0.83} & \cellcolor[RGB]{27,105,175}\textcolor{white}{0.96} & \cellcolor[RGB]{102,171,212}\textcolor{black}{0.78} & \cellcolor[RGB]{8,68,136}\textcolor{white}{0.96} \\
 & Batch & \cellcolor[RGB]{102,171,212}\textcolor{black}{0.26} & \cellcolor[RGB]{50,130,190}\textcolor{white}{0.54} & \cellcolor[RGB]{32,112,180}\textcolor{white}{0.91} & \cellcolor[RGB]{58,138,194}\textcolor{white}{0.92} & \cellcolor[RGB]{47,127,188}\textcolor{white}{0.89} & \cellcolor[RGB]{49,129,189}\textcolor{white}{0.94} & \cellcolor[RGB]{124,183,218}\textcolor{black}{0.76} & \cellcolor[RGB]{132,188,219}\textcolor{black}{0.85} \\
 & Batch, Few-shot & \cellcolor[RGB]{84,159,205}\textcolor{black}{0.25} & \cellcolor[RGB]{84,159,205}\textcolor{black}{0.50} & \cellcolor[RGB]{36,116,183}\textcolor{white}{0.90} & \cellcolor[RGB]{60,140,195}\textcolor{white}{0.92} & \cellcolor[RGB]{84,159,205}\textcolor{black}{0.85} & \cellcolor[RGB]{65,145,198}\textcolor{white}{0.93} & \cellcolor[RGB]{145,195,222}\textcolor{black}{0.73} & \cellcolor[RGB]{209,226,243}\textcolor{black}{0.80} \\
 & Description & \cellcolor[RGB]{8,48,107}\textcolor{white}{0.19} & \cellcolor[RGB]{8,69,138}\textcolor{white}{0.62} & \cellcolor[RGB]{8,55,118}\textcolor{white}{0.97} & \cellcolor[RGB]{8,48,107}\textcolor{white}{0.99} & \cellcolor[RGB]{10,84,158}\textcolor{white}{0.95} & \cellcolor[RGB]{8,65,132}\textcolor{white}{0.99} & \cellcolor[RGB]{30,109,178}\textcolor{white}{0.89} & \cellcolor[RGB]{8,65,132}\textcolor{white}{0.96} \\
 & Description, Few-shot & \cellcolor[RGB]{8,59,124}\textcolor{white}{0.19} & \cellcolor[RGB]{8,80,155}\textcolor{white}{0.61} & \cellcolor[RGB]{8,75,147}\textcolor{white}{0.95} & \cellcolor[RGB]{10,83,158}\textcolor{white}{0.97} & \cellcolor[RGB]{16,91,164}\textcolor{white}{0.94} & \cellcolor[RGB]{18,94,166}\textcolor{white}{0.97} & \cellcolor[RGB]{38,118,184}\textcolor{white}{0.88} & \cellcolor[RGB]{8,71,141}\textcolor{white}{0.96} \\
\midrule \multirow{8}{*}{Gemini} & Sampling & \cellcolor[RGB]{202,222,240}\textcolor{black}{0.31} & \cellcolor[RGB]{160,203,226}\textcolor{black}{0.43} & \cellcolor[RGB]{8,62,129}\textcolor{white}{0.96} & \cellcolor[RGB]{166,206,228}\textcolor{black}{0.86} & \cellcolor[RGB]{17,92,165}\textcolor{white}{0.94} & \cellcolor[RGB]{140,192,221}\textcolor{black}{0.89} & \cellcolor[RGB]{42,122,185}\textcolor{white}{0.87} & \cellcolor[RGB]{8,74,145}\textcolor{white}{0.95} \\
 & Sampling, Few-shot & \cellcolor[RGB]{146,196,222}\textcolor{black}{0.28} & \cellcolor[RGB]{220,233,246}\textcolor{black}{0.34} & \cellcolor[RGB]{238,245,252}\textcolor{black}{0.70} & \cellcolor[RGB]{222,235,247}\textcolor{black}{0.82} & \cellcolor[RGB]{196,218,238}\textcolor{black}{0.74} & \cellcolor[RGB]{247,251,255}\textcolor{black}{0.80} & \cellcolor[RGB]{164,204,227}\textcolor{black}{0.71} & \cellcolor[RGB]{228,239,249}\textcolor{black}{0.78} \\
 & Persona & \cellcolor[RGB]{50,130,190}\textcolor{white}{0.24} & \cellcolor[RGB]{77,153,202}\textcolor{black}{0.51} & \cellcolor[RGB]{19,95,167}\textcolor{white}{0.93} & \cellcolor[RGB]{14,88,162}\textcolor{white}{0.96} & \cellcolor[RGB]{28,106,176}\textcolor{white}{0.92} & \cellcolor[RGB]{16,91,164}\textcolor{white}{0.97} & \cellcolor[RGB]{91,163,208}\textcolor{black}{0.80} & \cellcolor[RGB]{8,53,115}\textcolor{white}{0.97} \\
 & Persona, Few-shot & \cellcolor[RGB]{50,130,190}\textcolor{white}{0.24} & \cellcolor[RGB]{205,223,241}\textcolor{black}{0.37} & \cellcolor[RGB]{234,243,251}\textcolor{black}{0.71} & \cellcolor[RGB]{195,218,238}\textcolor{black}{0.85} & \cellcolor[RGB]{180,211,233}\textcolor{black}{0.76} & \cellcolor[RGB]{204,223,241}\textcolor{black}{0.85} & \cellcolor[RGB]{145,195,222}\textcolor{black}{0.73} & \cellcolor[RGB]{124,183,218}\textcolor{black}{0.86} \\
 & Batch & \cellcolor[RGB]{72,150,200}\textcolor{white}{0.25} & \cellcolor[RGB]{77,153,202}\textcolor{black}{0.51} & \cellcolor[RGB]{8,74,145}\textcolor{white}{0.95} & \cellcolor[RGB]{8,58,122}\textcolor{white}{0.99} & \cellcolor[RGB]{61,141,196}\textcolor{white}{0.88} & \cellcolor[RGB]{8,75,147}\textcolor{white}{0.98} & \cellcolor[RGB]{129,186,219}\textcolor{black}{0.75} & \cellcolor[RGB]{8,77,150}\textcolor{white}{0.95} \\
 & Batch, Few-shot & \cellcolor[RGB]{65,145,198}\textcolor{white}{0.25} & \cellcolor[RGB]{48,128,189}\textcolor{white}{0.54} & \cellcolor[RGB]{8,60,125}\textcolor{white}{0.97} & \cellcolor[RGB]{8,70,139}\textcolor{white}{0.98} & \cellcolor[RGB]{26,104,174}\textcolor{white}{0.92} & \cellcolor[RGB]{10,84,158}\textcolor{white}{0.97} & \cellcolor[RGB]{84,159,205}\textcolor{black}{0.81} & \cellcolor[RGB]{47,127,188}\textcolor{white}{0.91} \\
 & Description & \cellcolor[RGB]{8,76,149}\textcolor{white}{0.20} & \cellcolor[RGB]{74,152,201}\textcolor{white}{0.51} & \cellcolor[RGB]{39,119,184}\textcolor{white}{0.90} & \cellcolor[RGB]{8,48,107}\textcolor{white}{0.99} & \cellcolor[RGB]{55,135,192}\textcolor{white}{0.88} & \cellcolor[RGB]{8,71,141}\textcolor{white}{0.98} & \cellcolor[RGB]{127,185,218}\textcolor{black}{0.75} & \cellcolor[RGB]{15,90,163}\textcolor{white}{0.94} \\
 & Description, Few-shot & \cellcolor[RGB]{14,89,162}\textcolor{white}{0.21} & \cellcolor[RGB]{181,212,233}\textcolor{black}{0.40} & \cellcolor[RGB]{184,213,234}\textcolor{black}{0.78} & \cellcolor[RGB]{43,123,186}\textcolor{white}{0.94} & \cellcolor[RGB]{157,202,225}\textcolor{black}{0.78} & \cellcolor[RGB]{53,133,191}\textcolor{white}{0.94} & \cellcolor[RGB]{149,197,223}\textcolor{black}{0.73} & \cellcolor[RGB]{56,136,193}\textcolor{white}{0.90} \\
\midrule \multirow{8}{*}{Mistral} & Sampling & \cellcolor[RGB]{137,190,220}\textcolor{black}{0.28} & \cellcolor[RGB]{222,235,247}\textcolor{black}{0.34} & \cellcolor[RGB]{8,62,129}\textcolor{white}{0.96} & \cellcolor[RGB]{9,82,157}\textcolor{white}{0.97} & \cellcolor[RGB]{101,170,212}\textcolor{black}{0.83} & \cellcolor[RGB]{18,94,166}\textcolor{white}{0.97} & \cellcolor[RGB]{169,207,229}\textcolor{black}{0.71} & \cellcolor[RGB]{8,53,115}\textcolor{white}{0.97} \\
 & Sampling, Few-shot & \cellcolor[RGB]{127,185,218}\textcolor{black}{0.27} & \cellcolor[RGB]{217,232,245}\textcolor{black}{0.35} & \cellcolor[RGB]{199,220,239}\textcolor{black}{0.76} & \cellcolor[RGB]{69,148,199}\textcolor{white}{0.92} & \cellcolor[RGB]{170,207,229}\textcolor{black}{0.77} & \cellcolor[RGB]{88,161,207}\textcolor{black}{0.91} & \cellcolor[RGB]{160,203,226}\textcolor{black}{0.72} & \cellcolor[RGB]{91,163,208}\textcolor{black}{0.88} \\
 & Persona & \cellcolor[RGB]{36,116,183}\textcolor{white}{0.23} & \cellcolor[RGB]{116,179,216}\textcolor{black}{0.47} & \cellcolor[RGB]{45,125,187}\textcolor{white}{0.90} & \cellcolor[RGB]{8,74,145}\textcolor{white}{0.97} & \cellcolor[RGB]{66,146,198}\textcolor{white}{0.87} & \cellcolor[RGB]{14,89,162}\textcolor{white}{0.97} & \cellcolor[RGB]{204,223,241}\textcolor{black}{0.65} & \cellcolor[RGB]{38,118,184}\textcolor{white}{0.92} \\
 & Persona, Few-shot & \cellcolor[RGB]{54,134,192}\textcolor{white}{0.24} & \cellcolor[RGB]{130,187,219}\textcolor{black}{0.45} & \cellcolor[RGB]{72,150,200}\textcolor{white}{0.87} & \cellcolor[RGB]{8,75,147}\textcolor{white}{0.97} & \cellcolor[RGB]{73,151,201}\textcolor{white}{0.86} & \cellcolor[RGB]{18,93,166}\textcolor{white}{0.97} & \cellcolor[RGB]{124,183,218}\textcolor{black}{0.76} & \cellcolor[RGB]{25,102,173}\textcolor{white}{0.93} \\
 & Batch & \cellcolor[RGB]{239,246,252}\textcolor{black}{0.34} & \cellcolor[RGB]{217,232,245}\textcolor{black}{0.35} & \cellcolor[RGB]{149,197,223}\textcolor{black}{0.81} & \cellcolor[RGB]{159,202,225}\textcolor{black}{0.87} & \cellcolor[RGB]{156,201,225}\textcolor{black}{0.79} & \cellcolor[RGB]{149,197,223}\textcolor{black}{0.88} & \cellcolor[RGB]{186,214,235}\textcolor{black}{0.68} & \cellcolor[RGB]{165,205,227}\textcolor{black}{0.83} \\
 & Batch, Few-shot & \cellcolor[RGB]{247,251,255}\textcolor{black}{0.35} & \cellcolor[RGB]{247,251,255}\textcolor{black}{0.30} & \cellcolor[RGB]{247,251,255}\textcolor{black}{0.69} & \cellcolor[RGB]{247,251,255}\textcolor{black}{0.79} & \cellcolor[RGB]{206,224,242}\textcolor{black}{0.73} & \cellcolor[RGB]{203,222,241}\textcolor{black}{0.85} & \cellcolor[RGB]{190,216,236}\textcolor{black}{0.67} & \cellcolor[RGB]{205,224,241}\textcolor{black}{0.80} \\
 & Description & \cellcolor[RGB]{8,61,127}\textcolor{white}{0.19} & \cellcolor[RGB]{29,108,177}\textcolor{white}{0.57} & \cellcolor[RGB]{19,95,167}\textcolor{white}{0.93} & \cellcolor[RGB]{23,100,171}\textcolor{white}{0.95} & \cellcolor[RGB]{34,114,182}\textcolor{white}{0.91} & \cellcolor[RGB]{28,107,176}\textcolor{white}{0.96} & \cellcolor[RGB]{95,166,209}\textcolor{black}{0.79} & \cellcolor[RGB]{37,117,183}\textcolor{white}{0.92} \\
 & Description, Few-shot & \cellcolor[RGB]{48,128,189}\textcolor{white}{0.24} & \cellcolor[RGB]{140,192,221}\textcolor{black}{0.45} & \cellcolor[RGB]{130,187,219}\textcolor{black}{0.82} & \cellcolor[RGB]{138,191,221}\textcolor{black}{0.88} & \cellcolor[RGB]{122,182,217}\textcolor{black}{0.81} & \cellcolor[RGB]{146,196,222}\textcolor{black}{0.88} & \cellcolor[RGB]{161,203,226}\textcolor{black}{0.72} & \cellcolor[RGB]{100,169,211}\textcolor{black}{0.87} \\
\midrule \multirow{5}{*}{Baseline} & Random & \cellcolor[RGB]{41,121,185}\textcolor{white}{0.23} & \cellcolor[RGB]{242,247,253}\textcolor{black}{0.31} & \cellcolor[RGB]{237,244,252}\textcolor{black}{0.71} & \cellcolor[RGB]{211,228,243}\textcolor{black}{0.83} & \cellcolor[RGB]{247,251,255}\textcolor{black}{0.65} & \cellcolor[RGB]{233,242,250}\textcolor{black}{0.82} & \cellcolor[RGB]{247,251,255}\textcolor{black}{0.55} & \cellcolor[RGB]{247,251,255}\textcolor{black}{0.76} \\
 & Popularity score &  & \cellcolor[RGB]{58,138,194}\textcolor{white}{0.53} & \cellcolor[RGB]{8,62,129}\textcolor{white}{0.96} & \cellcolor[RGB]{8,49,109}\textcolor{white}{0.99} & \cellcolor[RGB]{8,48,107}\textcolor{white}{1.00} & \cellcolor[RGB]{8,48,107}\textcolor{white}{1.00} & \cellcolor[RGB]{243,248,254}\textcolor{black}{0.56} & \cellcolor[RGB]{75,152,202}\textcolor{white}{0.89} \\
 & 5 real data & \cellcolor[RGB]{166,206,228}\textcolor{black}{0.29} & \cellcolor[RGB]{185,214,234}\textcolor{black}{0.40} & \cellcolor[RGB]{35,115,182}\textcolor{white}{0.91} & \cellcolor[RGB]{175,209,231}\textcolor{black}{0.86} & \cellcolor[RGB]{38,118,184}\textcolor{white}{0.91} & \cellcolor[RGB]{106,174,214}\textcolor{black}{0.90} & \cellcolor[RGB]{64,144,197}\textcolor{white}{0.83} & \cellcolor[RGB]{62,142,196}\textcolor{white}{0.90} \\
 & 15 real data & \cellcolor[RGB]{51,131,190}\textcolor{white}{0.24} & \cellcolor[RGB]{55,135,192}\textcolor{white}{0.53} & \cellcolor[RGB]{8,48,107}\textcolor{white}{0.98} & \cellcolor[RGB]{37,117,183}\textcolor{white}{0.94} & \cellcolor[RGB]{14,88,162}\textcolor{white}{0.95} & \cellcolor[RGB]{38,118,184}\textcolor{white}{0.95} & \cellcolor[RGB]{22,99,170}\textcolor{white}{0.91} & \cellcolor[RGB]{8,54,116}\textcolor{white}{0.97} \\
 & 30 real data & \cellcolor[RGB]{8,51,112}\textcolor{white}{0.19} & \cellcolor[RGB]{8,48,107}\textcolor{white}{0.65} & \cellcolor[RGB]{8,67,135}\textcolor{white}{0.96} & \cellcolor[RGB]{8,69,138}\textcolor{white}{0.98} & \cellcolor[RGB]{8,70,139}\textcolor{white}{0.97} & \cellcolor[RGB]{13,87,161}\textcolor{white}{0.97} & \cellcolor[RGB]{28,106,176}\textcolor{white}{0.90} & \cellcolor[RGB]{8,48,107}\textcolor{white}{0.97} \\
\bottomrule
\end{tabular}
\caption{Assortment optimization results for position-dependent rewards $r_p = 10/p$, averaged over 20 runs. Entries report the mean value for each model and method. Cell shading is column-wise, with darker indicating better performance within that metric. The model ``Gemini'' refers to Gemini 3 Flash, and ``Mistral'' refers to Mistral Large 3.}
\label{tab:assortment_results_10/i}
\end{table*}

\begin{table}[!htbp]
\footnotesize
\centering
\begin{tabular}{l l c c c c c c c c}
\toprule
 && \multirow{1}{*}{\textbf{Wasserstein}} & \multirow{1}{*}{\textbf{WorstCR}} & \multicolumn{6}{c}{\textbf{AverageCR} \textit{(higher is better)}} \\
 \cmidrule(lr){5-10}
 && \textit{(lower is}  & \textit{(higher  is} & \multicolumn{2}{c}{Unit} & \multicolumn{2}{c}{Random} & \multicolumn{2}{c}{Hard} \\
\textbf{Model} & \textbf{Method} & \textit{better)} &  \textit{better)} & $B=2$  & $B=5$  & $B=2$  & $B=5$  & $B=2$  & $B=5$ \\
\midrule \multirow{8}{*}{GPT-4o} & Sampling & \cellcolor[RGB]{47,127,188}\textcolor{white}{0.23} & \cellcolor[RGB]{52,132,191}\textcolor{white}{0.69} & \cellcolor[RGB]{8,72,142}\textcolor{white}{0.97} & \cellcolor[RGB]{38,118,184}\textcolor{white}{0.99} & \cellcolor[RGB]{25,102,173}\textcolor{white}{0.97} & \cellcolor[RGB]{58,138,194}\textcolor{white}{0.98} & \cellcolor[RGB]{171,208,230}\textcolor{black}{0.89} & \cellcolor[RGB]{14,88,162}\textcolor{white}{0.98} \\
 & Sampling, Few-shot & \cellcolor[RGB]{42,122,185}\textcolor{white}{0.23} & \cellcolor[RGB]{79,155,203}\textcolor{black}{0.65} & \cellcolor[RGB]{8,58,122}\textcolor{white}{0.97} & \cellcolor[RGB]{44,124,186}\textcolor{white}{0.98} & \cellcolor[RGB]{33,113,181}\textcolor{white}{0.96} & \cellcolor[RGB]{41,121,185}\textcolor{white}{0.98} & \cellcolor[RGB]{108,174,214}\textcolor{black}{0.92} & \cellcolor[RGB]{8,70,139}\textcolor{white}{0.99} \\
 & Persona & \cellcolor[RGB]{48,128,189}\textcolor{white}{0.24} & \cellcolor[RGB]{53,133,191}\textcolor{white}{0.69} & \cellcolor[RGB]{8,72,142}\textcolor{white}{0.97} & \cellcolor[RGB]{8,54,116}\textcolor{white}{1.00} & \cellcolor[RGB]{14,89,162}\textcolor{white}{0.98} & \cellcolor[RGB]{8,65,132}\textcolor{white}{1.00} & \cellcolor[RGB]{184,213,234}\textcolor{black}{0.88} & \cellcolor[RGB]{8,50,110}\textcolor{white}{0.99} \\
 & Persona, Few-shot & \cellcolor[RGB]{28,106,176}\textcolor{white}{0.22} & \cellcolor[RGB]{75,152,202}\textcolor{white}{0.66} & \cellcolor[RGB]{8,58,122}\textcolor{white}{0.97} & \cellcolor[RGB]{8,69,138}\textcolor{white}{1.00} & \cellcolor[RGB]{31,110,179}\textcolor{white}{0.96} & \cellcolor[RGB]{8,76,149}\textcolor{white}{0.99} & \cellcolor[RGB]{74,152,201}\textcolor{white}{0.93} & \cellcolor[RGB]{8,48,107}\textcolor{white}{0.99} \\
 & Batch & \cellcolor[RGB]{234,242,251}\textcolor{black}{0.34} & \cellcolor[RGB]{174,209,231}\textcolor{black}{0.57} & \cellcolor[RGB]{100,169,211}\textcolor{black}{0.92} & \cellcolor[RGB]{148,196,223}\textcolor{black}{0.97} & \cellcolor[RGB]{161,203,226}\textcolor{black}{0.91} & \cellcolor[RGB]{109,175,215}\textcolor{black}{0.97} & \cellcolor[RGB]{203,222,241}\textcolor{black}{0.87} & \cellcolor[RGB]{29,108,177}\textcolor{white}{0.98} \\
 & Batch, Few-shot & \cellcolor[RGB]{238,245,252}\textcolor{black}{0.34} & \cellcolor[RGB]{170,207,229}\textcolor{black}{0.57} & \cellcolor[RGB]{173,208,230}\textcolor{black}{0.90} & \cellcolor[RGB]{205,224,241}\textcolor{black}{0.96} & \cellcolor[RGB]{156,201,225}\textcolor{black}{0.91} & \cellcolor[RGB]{154,200,224}\textcolor{black}{0.97} & \cellcolor[RGB]{231,241,250}\textcolor{black}{0.84} & \cellcolor[RGB]{146,196,222}\textcolor{black}{0.95} \\
 & Description & \cellcolor[RGB]{8,66,133}\textcolor{white}{0.20} & \cellcolor[RGB]{37,117,183}\textcolor{white}{0.71} & \cellcolor[RGB]{28,106,176}\textcolor{white}{0.95} & \cellcolor[RGB]{38,118,184}\textcolor{white}{0.99} & \cellcolor[RGB]{41,121,185}\textcolor{white}{0.96} & \cellcolor[RGB]{34,114,182}\textcolor{white}{0.99} & \cellcolor[RGB]{42,122,185}\textcolor{white}{0.95} & \cellcolor[RGB]{15,90,163}\textcolor{white}{0.98} \\
 & Description, Few-shot & \cellcolor[RGB]{8,61,127}\textcolor{white}{0.19} & \cellcolor[RGB]{92,164,208}\textcolor{black}{0.64} & \cellcolor[RGB]{12,86,160}\textcolor{white}{0.96} & \cellcolor[RGB]{28,106,176}\textcolor{white}{0.99} & \cellcolor[RGB]{27,105,175}\textcolor{white}{0.97} & \cellcolor[RGB]{28,106,176}\textcolor{white}{0.99} & \cellcolor[RGB]{33,113,181}\textcolor{white}{0.96} & \cellcolor[RGB]{42,122,185}\textcolor{white}{0.97} \\
\midrule \multirow{8}{*}{GPT-5-mini} & Sampling & \cellcolor[RGB]{193,217,237}\textcolor{black}{0.30} & \cellcolor[RGB]{132,188,219}\textcolor{black}{0.60} & \cellcolor[RGB]{8,72,142}\textcolor{white}{0.97} & \cellcolor[RGB]{201,221,240}\textcolor{black}{0.96} & \cellcolor[RGB]{15,90,163}\textcolor{white}{0.98} & \cellcolor[RGB]{146,196,222}\textcolor{black}{0.97} & \cellcolor[RGB]{8,48,107}\textcolor{white}{1.00} & \cellcolor[RGB]{10,84,158}\textcolor{white}{0.98} \\
 & Sampling, Few-shot & \cellcolor[RGB]{99,168,211}\textcolor{black}{0.26} & \cellcolor[RGB]{122,182,217}\textcolor{black}{0.61} & \cellcolor[RGB]{21,98,169}\textcolor{white}{0.96} & \cellcolor[RGB]{23,100,171}\textcolor{white}{0.99} & \cellcolor[RGB]{14,89,162}\textcolor{white}{0.98} & \cellcolor[RGB]{31,110,179}\textcolor{white}{0.99} & \cellcolor[RGB]{12,86,160}\textcolor{white}{0.98} & \cellcolor[RGB]{14,88,162}\textcolor{white}{0.98} \\
 & Persona & \cellcolor[RGB]{22,99,170}\textcolor{white}{0.22} & \cellcolor[RGB]{64,144,197}\textcolor{white}{0.67} & \cellcolor[RGB]{65,145,198}\textcolor{white}{0.93} & \cellcolor[RGB]{17,92,165}\textcolor{white}{0.99} & \cellcolor[RGB]{87,160,206}\textcolor{black}{0.93} & \cellcolor[RGB]{21,98,169}\textcolor{white}{0.99} & \cellcolor[RGB]{171,208,230}\textcolor{black}{0.89} & \cellcolor[RGB]{8,80,155}\textcolor{white}{0.99} \\
 & Persona, Few-shot & \cellcolor[RGB]{25,102,173}\textcolor{white}{0.22} & \cellcolor[RGB]{53,133,191}\textcolor{white}{0.69} & \cellcolor[RGB]{12,86,160}\textcolor{white}{0.96} & \cellcolor[RGB]{12,86,160}\textcolor{white}{0.99} & \cellcolor[RGB]{47,127,188}\textcolor{white}{0.96} & \cellcolor[RGB]{28,107,176}\textcolor{white}{0.99} & \cellcolor[RGB]{121,181,217}\textcolor{black}{0.91} & \cellcolor[RGB]{14,88,162}\textcolor{white}{0.98} \\
 & Batch & \cellcolor[RGB]{102,171,212}\textcolor{black}{0.26} & \cellcolor[RGB]{101,170,212}\textcolor{black}{0.63} & \cellcolor[RGB]{23,100,171}\textcolor{white}{0.95} & \cellcolor[RGB]{92,164,208}\textcolor{black}{0.98} & \cellcolor[RGB]{43,123,186}\textcolor{white}{0.96} & \cellcolor[RGB]{81,156,204}\textcolor{black}{0.98} & \cellcolor[RGB]{60,140,195}\textcolor{white}{0.94} & \cellcolor[RGB]{48,128,189}\textcolor{white}{0.97} \\
 & Batch, Few-shot & \cellcolor[RGB]{84,159,205}\textcolor{black}{0.25} & \cellcolor[RGB]{88,161,207}\textcolor{black}{0.65} & \cellcolor[RGB]{59,139,194}\textcolor{white}{0.94} & \cellcolor[RGB]{87,160,206}\textcolor{black}{0.98} & \cellcolor[RGB]{70,149,200}\textcolor{white}{0.94} & \cellcolor[RGB]{58,138,194}\textcolor{white}{0.98} & \cellcolor[RGB]{124,183,218}\textcolor{black}{0.91} & \cellcolor[RGB]{55,135,192}\textcolor{white}{0.97} \\
 & Description & \cellcolor[RGB]{8,48,107}\textcolor{white}{0.19} & \cellcolor[RGB]{50,130,190}\textcolor{white}{0.69} & \cellcolor[RGB]{9,82,157}\textcolor{white}{0.96} & \cellcolor[RGB]{8,68,136}\textcolor{white}{1.00} & \cellcolor[RGB]{29,108,177}\textcolor{white}{0.97} & \cellcolor[RGB]{8,75,147}\textcolor{white}{0.99} & \cellcolor[RGB]{20,96,168}\textcolor{white}{0.97} & \cellcolor[RGB]{8,69,138}\textcolor{white}{0.99} \\
 & Description, Few-shot & \cellcolor[RGB]{8,59,124}\textcolor{white}{0.19} & \cellcolor[RGB]{62,142,196}\textcolor{white}{0.67} & \cellcolor[RGB]{14,89,162}\textcolor{white}{0.96} & \cellcolor[RGB]{22,99,170}\textcolor{white}{0.99} & \cellcolor[RGB]{28,106,176}\textcolor{white}{0.97} & \cellcolor[RGB]{21,98,169}\textcolor{white}{0.99} & \cellcolor[RGB]{18,94,166}\textcolor{white}{0.97} & \cellcolor[RGB]{16,91,164}\textcolor{white}{0.98} \\
\midrule \multirow{8}{*}{Gemini} & Sampling & \cellcolor[RGB]{202,222,240}\textcolor{black}{0.31} & \cellcolor[RGB]{132,188,219}\textcolor{black}{0.60} & \cellcolor[RGB]{8,72,142}\textcolor{white}{0.97} & \cellcolor[RGB]{201,221,240}\textcolor{black}{0.96} & \cellcolor[RGB]{21,98,169}\textcolor{white}{0.97} & \cellcolor[RGB]{163,204,227}\textcolor{black}{0.97} & \cellcolor[RGB]{8,48,107}\textcolor{white}{1.00} & \cellcolor[RGB]{52,132,191}\textcolor{white}{0.97} \\
 & Sampling, Few-shot & \cellcolor[RGB]{146,196,222}\textcolor{black}{0.28} & \cellcolor[RGB]{114,178,216}\textcolor{black}{0.62} & \cellcolor[RGB]{178,210,232}\textcolor{black}{0.89} & \cellcolor[RGB]{229,239,249}\textcolor{black}{0.95} & \cellcolor[RGB]{160,203,226}\textcolor{black}{0.91} & \cellcolor[RGB]{247,251,255}\textcolor{black}{0.95} & \cellcolor[RGB]{246,250,255}\textcolor{black}{0.83} & \cellcolor[RGB]{247,251,255}\textcolor{black}{0.92} \\
 & Persona & \cellcolor[RGB]{50,130,190}\textcolor{white}{0.24} & \cellcolor[RGB]{92,164,208}\textcolor{black}{0.64} & \cellcolor[RGB]{8,70,139}\textcolor{white}{0.97} & \cellcolor[RGB]{24,101,172}\textcolor{white}{0.99} & \cellcolor[RGB]{14,89,162}\textcolor{white}{0.98} & \cellcolor[RGB]{21,98,169}\textcolor{white}{0.99} & \cellcolor[RGB]{8,48,107}\textcolor{white}{1.00} & \cellcolor[RGB]{8,66,133}\textcolor{white}{0.99} \\
 & Persona, Few-shot & \cellcolor[RGB]{50,130,190}\textcolor{white}{0.24} & \cellcolor[RGB]{72,150,200}\textcolor{white}{0.66} & \cellcolor[RGB]{148,196,223}\textcolor{black}{0.90} & \cellcolor[RGB]{181,212,233}\textcolor{black}{0.96} & \cellcolor[RGB]{86,160,206}\textcolor{black}{0.94} & \cellcolor[RGB]{178,210,232}\textcolor{black}{0.96} & \cellcolor[RGB]{82,157,204}\textcolor{black}{0.93} & \cellcolor[RGB]{87,160,206}\textcolor{black}{0.96} \\
 & Batch & \cellcolor[RGB]{72,150,200}\textcolor{white}{0.25} & \cellcolor[RGB]{64,144,197}\textcolor{white}{0.67} & \cellcolor[RGB]{16,91,164}\textcolor{white}{0.96} & \cellcolor[RGB]{8,74,145}\textcolor{white}{0.99} & \cellcolor[RGB]{52,132,191}\textcolor{white}{0.95} & \cellcolor[RGB]{10,83,158}\textcolor{white}{0.99} & \cellcolor[RGB]{129,186,219}\textcolor{black}{0.91} & \cellcolor[RGB]{8,70,139}\textcolor{white}{0.99} \\
 & Batch, Few-shot & \cellcolor[RGB]{65,145,198}\textcolor{white}{0.25} & \cellcolor[RGB]{74,152,201}\textcolor{white}{0.66} & \cellcolor[RGB]{8,72,142}\textcolor{white}{0.97} & \cellcolor[RGB]{8,73,144}\textcolor{white}{0.99} & \cellcolor[RGB]{27,105,175}\textcolor{white}{0.97} & \cellcolor[RGB]{8,78,152}\textcolor{white}{0.99} & \cellcolor[RGB]{92,164,208}\textcolor{black}{0.92} & \cellcolor[RGB]{8,72,142}\textcolor{white}{0.99} \\
 & Description & \cellcolor[RGB]{8,76,149}\textcolor{white}{0.20} & \cellcolor[RGB]{96,167,210}\textcolor{black}{0.64} & \cellcolor[RGB]{52,132,191}\textcolor{white}{0.94} & \cellcolor[RGB]{8,77,150}\textcolor{white}{0.99} & \cellcolor[RGB]{75,152,202}\textcolor{white}{0.94} & \cellcolor[RGB]{11,85,159}\textcolor{white}{0.99} & \cellcolor[RGB]{117,180,216}\textcolor{black}{0.91} & \cellcolor[RGB]{8,72,142}\textcolor{white}{0.99} \\
 & Description, Few-shot & \cellcolor[RGB]{14,89,162}\textcolor{white}{0.21} & \cellcolor[RGB]{111,176,215}\textcolor{black}{0.62} & \cellcolor[RGB]{135,189,220}\textcolor{black}{0.91} & \cellcolor[RGB]{47,127,188}\textcolor{white}{0.98} & \cellcolor[RGB]{114,178,216}\textcolor{black}{0.92} & \cellcolor[RGB]{49,129,189}\textcolor{white}{0.98} & \cellcolor[RGB]{206,224,242}\textcolor{black}{0.87} & \cellcolor[RGB]{20,96,168}\textcolor{white}{0.98} \\
\midrule \multirow{8}{*}{Mistral} & Sampling & \cellcolor[RGB]{137,190,220}\textcolor{black}{0.28} & \cellcolor[RGB]{236,244,251}\textcolor{black}{0.47} & \cellcolor[RGB]{8,72,142}\textcolor{white}{0.97} & \cellcolor[RGB]{14,88,162}\textcolor{white}{0.99} & \cellcolor[RGB]{88,161,207}\textcolor{black}{0.93} & \cellcolor[RGB]{18,94,166}\textcolor{white}{0.99} & \cellcolor[RGB]{175,209,231}\textcolor{black}{0.88} & \cellcolor[RGB]{8,67,135}\textcolor{white}{0.99} \\
 & Sampling, Few-shot & \cellcolor[RGB]{127,185,218}\textcolor{black}{0.27} & \cellcolor[RGB]{233,242,250}\textcolor{black}{0.48} & \cellcolor[RGB]{46,126,188}\textcolor{white}{0.94} & \cellcolor[RGB]{34,114,182}\textcolor{white}{0.99} & \cellcolor[RGB]{81,156,204}\textcolor{black}{0.94} & \cellcolor[RGB]{52,132,191}\textcolor{white}{0.98} & \cellcolor[RGB]{200,220,240}\textcolor{black}{0.87} & \cellcolor[RGB]{19,95,167}\textcolor{white}{0.98} \\
 & Persona & \cellcolor[RGB]{36,116,183}\textcolor{white}{0.23} & \cellcolor[RGB]{159,202,225}\textcolor{black}{0.58} & \cellcolor[RGB]{8,78,152}\textcolor{white}{0.97} & \cellcolor[RGB]{8,78,152}\textcolor{white}{0.99} & \cellcolor[RGB]{33,113,181}\textcolor{white}{0.96} & \cellcolor[RGB]{13,87,161}\textcolor{white}{0.99} & \cellcolor[RGB]{230,240,249}\textcolor{black}{0.85} & \cellcolor[RGB]{8,73,144}\textcolor{white}{0.99} \\
 & Persona, Few-shot & \cellcolor[RGB]{54,134,192}\textcolor{white}{0.24} & \cellcolor[RGB]{183,212,234}\textcolor{black}{0.56} & \cellcolor[RGB]{46,126,188}\textcolor{white}{0.94} & \cellcolor[RGB]{10,84,158}\textcolor{white}{0.99} & \cellcolor[RGB]{55,135,192}\textcolor{white}{0.95} & \cellcolor[RGB]{16,91,164}\textcolor{white}{0.99} & \cellcolor[RGB]{201,221,240}\textcolor{black}{0.87} & \cellcolor[RGB]{8,58,122}\textcolor{white}{0.99} \\
 & Batch & \cellcolor[RGB]{239,246,252}\textcolor{black}{0.34} & \cellcolor[RGB]{231,241,250}\textcolor{black}{0.48} & \cellcolor[RGB]{113,177,215}\textcolor{black}{0.92} & \cellcolor[RGB]{196,218,238}\textcolor{black}{0.96} & \cellcolor[RGB]{124,183,218}\textcolor{black}{0.92} & \cellcolor[RGB]{163,204,227}\textcolor{black}{0.97} & \cellcolor[RGB]{181,212,233}\textcolor{black}{0.88} & \cellcolor[RGB]{72,150,200}\textcolor{white}{0.96} \\
 & Batch, Few-shot & \cellcolor[RGB]{247,251,255}\textcolor{black}{0.35} & \cellcolor[RGB]{238,245,252}\textcolor{black}{0.47} & \cellcolor[RGB]{205,223,241}\textcolor{black}{0.88} & \cellcolor[RGB]{247,251,255}\textcolor{black}{0.95} & \cellcolor[RGB]{166,206,228}\textcolor{black}{0.90} & \cellcolor[RGB]{171,208,230}\textcolor{black}{0.96} & \cellcolor[RGB]{239,246,252}\textcolor{black}{0.84} & \cellcolor[RGB]{200,220,240}\textcolor{black}{0.94} \\
 & Description & \cellcolor[RGB]{8,61,127}\textcolor{white}{0.19} & \cellcolor[RGB]{62,142,196}\textcolor{white}{0.67} & \cellcolor[RGB]{23,100,171}\textcolor{white}{0.95} & \cellcolor[RGB]{28,107,176}\textcolor{white}{0.99} & \cellcolor[RGB]{38,118,184}\textcolor{white}{0.96} & \cellcolor[RGB]{29,108,177}\textcolor{white}{0.99} & \cellcolor[RGB]{57,137,193}\textcolor{white}{0.94} & \cellcolor[RGB]{14,88,162}\textcolor{white}{0.98} \\
 & Description, Few-shot & \cellcolor[RGB]{48,128,189}\textcolor{white}{0.24} & \cellcolor[RGB]{202,222,240}\textcolor{black}{0.53} & \cellcolor[RGB]{153,199,224}\textcolor{black}{0.90} & \cellcolor[RGB]{186,214,235}\textcolor{black}{0.96} & \cellcolor[RGB]{151,198,223}\textcolor{black}{0.91} & \cellcolor[RGB]{173,208,230}\textcolor{black}{0.96} & \cellcolor[RGB]{217,232,245}\textcolor{black}{0.86} & \cellcolor[RGB]{95,166,209}\textcolor{black}{0.96} \\
\midrule \multirow{5}{*}{Baseline} & Random & \cellcolor[RGB]{41,121,185}\textcolor{white}{0.23} & \cellcolor[RGB]{247,251,255}\textcolor{black}{0.46} & \cellcolor[RGB]{247,251,255}\textcolor{black}{0.85} & \cellcolor[RGB]{215,230,245}\textcolor{black}{0.96} & \cellcolor[RGB]{247,251,255}\textcolor{black}{0.85} & \cellcolor[RGB]{210,227,243}\textcolor{black}{0.96} & \cellcolor[RGB]{247,251,255}\textcolor{black}{0.83} & \cellcolor[RGB]{194,217,238}\textcolor{black}{0.94} \\
 & Popularity score &  & \cellcolor[RGB]{38,118,184}\textcolor{white}{0.70} & \cellcolor[RGB]{8,72,142}\textcolor{white}{0.97} & \cellcolor[RGB]{8,48,107}\textcolor{white}{1.00} & \cellcolor[RGB]{8,48,107}\textcolor{white}{1.00} & \cellcolor[RGB]{8,48,107}\textcolor{white}{1.00} & \cellcolor[RGB]{205,223,241}\textcolor{black}{0.87} & \cellcolor[RGB]{12,86,160}\textcolor{white}{0.98} \\
 & 5 real data & \cellcolor[RGB]{166,206,228}\textcolor{black}{0.29} & \cellcolor[RGB]{129,186,219}\textcolor{black}{0.61} & \cellcolor[RGB]{44,124,186}\textcolor{white}{0.94} & \cellcolor[RGB]{193,217,237}\textcolor{black}{0.96} & \cellcolor[RGB]{54,134,192}\textcolor{white}{0.95} & \cellcolor[RGB]{106,174,214}\textcolor{black}{0.97} & \cellcolor[RGB]{74,152,201}\textcolor{white}{0.93} & \cellcolor[RGB]{51,131,190}\textcolor{white}{0.97} \\
 & 15 real data & \cellcolor[RGB]{51,131,190}\textcolor{white}{0.24} & \cellcolor[RGB]{35,115,182}\textcolor{white}{0.71} & \cellcolor[RGB]{8,75,147}\textcolor{white}{0.97} & \cellcolor[RGB]{45,125,187}\textcolor{white}{0.98} & \cellcolor[RGB]{15,90,163}\textcolor{white}{0.98} & \cellcolor[RGB]{47,127,188}\textcolor{white}{0.98} & \cellcolor[RGB]{8,62,129}\textcolor{white}{0.99} & \cellcolor[RGB]{8,78,152}\textcolor{white}{0.99} \\
 & 30 real data & \cellcolor[RGB]{8,51,112}\textcolor{white}{0.19} & \cellcolor[RGB]{8,48,107}\textcolor{white}{0.80} & \cellcolor[RGB]{8,48,107}\textcolor{white}{0.98} & \cellcolor[RGB]{17,92,165}\textcolor{white}{0.99} & \cellcolor[RGB]{8,72,142}\textcolor{white}{0.99} & \cellcolor[RGB]{18,93,166}\textcolor{white}{0.99} & \cellcolor[RGB]{11,85,159}\textcolor{white}{0.98} & \cellcolor[RGB]{8,58,122}\textcolor{white}{0.99} \\
\bottomrule
\end{tabular}
\caption{Assortment optimization results for position-dependent rewards $r_p=10-(p-1)$, averaged over 20 runs. Entries report the mean value for each model and method. Cell shading is column-wise, with darker indicating better performance within that metric. The model ``Gemini'' refers to Gemini 3 Flash, and ``Mistral'' refers to Mistral Large 3.}
\label{tab:assortment_results_10-i}
\end{table}

\begin{table*}[!htbp]
\footnotesize
\centering
\begin{tabular}{l l c c c c c c c c}
\toprule
 && \multirow{1}{*}{\textbf{Wasserstein}} & \multirow{1}{*}{\textbf{WorstCR}} & \multicolumn{6}{c}{\textbf{AverageCR} \textit{(higher is better)}} \\
 \cmidrule(lr){5-10}
 && \textit{(lower is}  & \textit{(higher  is} & \multicolumn{2}{c}{Unit} & \multicolumn{2}{c}{Random} & \multicolumn{2}{c}{Hard} \\
\textbf{Model} & \textbf{Method} & \textit{better)} &  \textit{better)} & $B=2$  & $B=5$  & $B=2$  & $B=5$  & $B=2$  & $B=5$ \\
\midrule \multirow{8}{*}{GPT-4o} & Sampling & \cellcolor[RGB]{47,127,188}\textcolor{white}{0.23} & \cellcolor[RGB]{79,155,203}\textcolor{black}{0.75} & \cellcolor[RGB]{21,97,169}\textcolor{white}{0.98} & \cellcolor[RGB]{22,99,170}\textcolor{white}{1.00} & \cellcolor[RGB]{23,100,171}\textcolor{white}{0.98} & \cellcolor[RGB]{42,122,185}\textcolor{white}{1.00} & \cellcolor[RGB]{105,173,213}\textcolor{black}{0.96} & \cellcolor[RGB]{8,64,130}\textcolor{white}{1.00} \\
 & Sampling, Few-shot & \cellcolor[RGB]{42,122,185}\textcolor{white}{0.23} & \cellcolor[RGB]{96,167,210}\textcolor{black}{0.74} & \cellcolor[RGB]{8,74,145}\textcolor{white}{0.98} & \cellcolor[RGB]{41,121,185}\textcolor{white}{1.00} & \cellcolor[RGB]{30,109,178}\textcolor{white}{0.98} & \cellcolor[RGB]{37,117,183}\textcolor{white}{1.00} & \cellcolor[RGB]{108,174,214}\textcolor{black}{0.96} & \cellcolor[RGB]{8,57,121}\textcolor{white}{1.00} \\
 & Persona & \cellcolor[RGB]{48,128,189}\textcolor{white}{0.24} & \cellcolor[RGB]{91,163,208}\textcolor{black}{0.74} & \cellcolor[RGB]{21,97,169}\textcolor{white}{0.98} & \cellcolor[RGB]{8,62,129}\textcolor{white}{1.00} & \cellcolor[RGB]{24,101,172}\textcolor{white}{0.98} & \cellcolor[RGB]{8,70,139}\textcolor{white}{1.00} & \cellcolor[RGB]{124,183,218}\textcolor{black}{0.95} & \cellcolor[RGB]{8,58,122}\textcolor{white}{1.00} \\
 & Persona, Few-shot & \cellcolor[RGB]{28,106,176}\textcolor{white}{0.22} & \cellcolor[RGB]{97,167,210}\textcolor{black}{0.74} & \cellcolor[RGB]{14,88,162}\textcolor{white}{0.98} & \cellcolor[RGB]{8,74,145}\textcolor{white}{1.00} & \cellcolor[RGB]{30,109,178}\textcolor{white}{0.98} & \cellcolor[RGB]{10,83,158}\textcolor{white}{1.00} & \cellcolor[RGB]{90,162,207}\textcolor{black}{0.96} & \cellcolor[RGB]{8,65,132}\textcolor{white}{1.00} \\
 & Batch & \cellcolor[RGB]{234,242,251}\textcolor{black}{0.34} & \cellcolor[RGB]{151,198,223}\textcolor{black}{0.70} & \cellcolor[RGB]{161,203,226}\textcolor{black}{0.96} & \cellcolor[RGB]{163,204,227}\textcolor{black}{0.99} & \cellcolor[RGB]{159,202,225}\textcolor{black}{0.95} & \cellcolor[RGB]{138,191,221}\textcolor{black}{0.99} & \cellcolor[RGB]{210,227,243}\textcolor{black}{0.93} & \cellcolor[RGB]{29,108,177}\textcolor{white}{0.99} \\
 & Batch, Few-shot & \cellcolor[RGB]{238,245,252}\textcolor{black}{0.34} & \cellcolor[RGB]{138,191,221}\textcolor{black}{0.70} & \cellcolor[RGB]{174,209,231}\textcolor{black}{0.95} & \cellcolor[RGB]{184,213,234}\textcolor{black}{0.99} & \cellcolor[RGB]{114,178,216}\textcolor{black}{0.96} & \cellcolor[RGB]{149,197,223}\textcolor{black}{0.99} & \cellcolor[RGB]{237,244,252}\textcolor{black}{0.92} & \cellcolor[RGB]{114,178,216}\textcolor{black}{0.99} \\
 & Description & \cellcolor[RGB]{8,66,133}\textcolor{white}{0.20} & \cellcolor[RGB]{59,139,194}\textcolor{white}{0.77} & \cellcolor[RGB]{84,159,205}\textcolor{black}{0.97} & \cellcolor[RGB]{58,138,194}\textcolor{white}{1.00} & \cellcolor[RGB]{59,139,194}\textcolor{white}{0.97} & \cellcolor[RGB]{57,137,193}\textcolor{white}{1.00} & \cellcolor[RGB]{72,150,200}\textcolor{white}{0.97} & \cellcolor[RGB]{17,92,165}\textcolor{white}{0.99} \\
 & Description, Few-shot & \cellcolor[RGB]{8,61,127}\textcolor{white}{0.19} & \cellcolor[RGB]{96,167,210}\textcolor{black}{0.74} & \cellcolor[RGB]{28,106,176}\textcolor{white}{0.98} & \cellcolor[RGB]{26,104,174}\textcolor{white}{1.00} & \cellcolor[RGB]{32,111,180}\textcolor{white}{0.98} & \cellcolor[RGB]{28,106,176}\textcolor{white}{1.00} & \cellcolor[RGB]{57,137,193}\textcolor{white}{0.97} & \cellcolor[RGB]{25,103,173}\textcolor{white}{0.99} \\
\midrule \multirow{8}{*}{GPT-5-mini} & Sampling & \cellcolor[RGB]{193,217,237}\textcolor{black}{0.30} & \cellcolor[RGB]{137,190,220}\textcolor{black}{0.71} & \cellcolor[RGB]{21,97,169}\textcolor{white}{0.98} & \cellcolor[RGB]{216,231,245}\textcolor{black}{0.99} & \cellcolor[RGB]{29,108,177}\textcolor{white}{0.98} & \cellcolor[RGB]{175,209,231}\textcolor{black}{0.99} & \cellcolor[RGB]{8,48,107}\textcolor{white}{1.00} & \cellcolor[RGB]{8,78,152}\textcolor{white}{1.00} \\
 & Sampling, Few-shot & \cellcolor[RGB]{99,168,211}\textcolor{black}{0.26} & \cellcolor[RGB]{125,184,218}\textcolor{black}{0.71} & \cellcolor[RGB]{35,115,182}\textcolor{white}{0.97} & \cellcolor[RGB]{34,114,182}\textcolor{white}{1.00} & \cellcolor[RGB]{16,91,164}\textcolor{white}{0.99} & \cellcolor[RGB]{50,130,190}\textcolor{white}{1.00} & \cellcolor[RGB]{13,87,161}\textcolor{white}{0.99} & \cellcolor[RGB]{28,106,176}\textcolor{white}{0.99} \\
 & Persona & \cellcolor[RGB]{22,99,170}\textcolor{white}{0.22} & \cellcolor[RGB]{60,140,195}\textcolor{white}{0.77} & \cellcolor[RGB]{32,112,180}\textcolor{white}{0.97} & \cellcolor[RGB]{18,93,166}\textcolor{white}{1.00} & \cellcolor[RGB]{39,119,184}\textcolor{white}{0.98} & \cellcolor[RGB]{18,94,166}\textcolor{white}{1.00} & \cellcolor[RGB]{105,173,213}\textcolor{black}{0.96} & \cellcolor[RGB]{32,111,180}\textcolor{white}{0.99} \\
 & Persona, Few-shot & \cellcolor[RGB]{25,102,173}\textcolor{white}{0.22} & \cellcolor[RGB]{72,150,200}\textcolor{white}{0.76} & \cellcolor[RGB]{41,121,185}\textcolor{white}{0.97} & \cellcolor[RGB]{28,106,176}\textcolor{white}{1.00} & \cellcolor[RGB]{32,112,180}\textcolor{white}{0.98} & \cellcolor[RGB]{32,112,180}\textcolor{white}{1.00} & \cellcolor[RGB]{105,173,213}\textcolor{black}{0.96} & \cellcolor[RGB]{8,77,150}\textcolor{white}{1.00} \\
 & Batch & \cellcolor[RGB]{102,171,212}\textcolor{black}{0.26} & \cellcolor[RGB]{140,192,221}\textcolor{black}{0.70} & \cellcolor[RGB]{62,142,196}\textcolor{white}{0.97} & \cellcolor[RGB]{63,143,197}\textcolor{white}{1.00} & \cellcolor[RGB]{54,134,192}\textcolor{white}{0.98} & \cellcolor[RGB]{69,148,199}\textcolor{white}{0.99} & \cellcolor[RGB]{59,139,194}\textcolor{white}{0.97} & \cellcolor[RGB]{34,114,182}\textcolor{white}{0.99} \\
 & Batch, Few-shot & \cellcolor[RGB]{84,159,205}\textcolor{black}{0.25} & \cellcolor[RGB]{114,178,216}\textcolor{black}{0.72} & \cellcolor[RGB]{62,142,196}\textcolor{white}{0.97} & \cellcolor[RGB]{69,148,199}\textcolor{white}{0.99} & \cellcolor[RGB]{81,156,204}\textcolor{black}{0.97} & \cellcolor[RGB]{50,130,190}\textcolor{white}{1.00} & \cellcolor[RGB]{108,174,214}\textcolor{black}{0.96} & \cellcolor[RGB]{18,93,166}\textcolor{white}{0.99} \\
 & Description & \cellcolor[RGB]{8,48,107}\textcolor{white}{0.19} & \cellcolor[RGB]{79,155,203}\textcolor{black}{0.75} & \cellcolor[RGB]{47,127,188}\textcolor{white}{0.97} & \cellcolor[RGB]{11,85,159}\textcolor{white}{1.00} & \cellcolor[RGB]{51,131,190}\textcolor{white}{0.98} & \cellcolor[RGB]{18,94,166}\textcolor{white}{1.00} & \cellcolor[RGB]{93,165,209}\textcolor{black}{0.96} & \cellcolor[RGB]{8,79,153}\textcolor{white}{1.00} \\
 & Description, Few-shot & \cellcolor[RGB]{8,59,124}\textcolor{white}{0.19} & \cellcolor[RGB]{91,163,208}\textcolor{black}{0.74} & \cellcolor[RGB]{32,112,180}\textcolor{white}{0.97} & \cellcolor[RGB]{26,104,174}\textcolor{white}{1.00} & \cellcolor[RGB]{41,121,185}\textcolor{white}{0.98} & \cellcolor[RGB]{27,105,175}\textcolor{white}{1.00} & \cellcolor[RGB]{66,146,198}\textcolor{white}{0.97} & \cellcolor[RGB]{17,92,165}\textcolor{white}{0.99} \\
\midrule \multirow{8}{*}{Gemini} & Sampling & \cellcolor[RGB]{202,222,240}\textcolor{black}{0.31} & \cellcolor[RGB]{137,190,220}\textcolor{black}{0.71} & \cellcolor[RGB]{21,97,169}\textcolor{white}{0.98} & \cellcolor[RGB]{216,231,245}\textcolor{black}{0.99} & \cellcolor[RGB]{27,105,175}\textcolor{white}{0.98} & \cellcolor[RGB]{200,220,240}\textcolor{black}{0.99} & \cellcolor[RGB]{8,48,107}\textcolor{white}{1.00} & \cellcolor[RGB]{90,162,207}\textcolor{black}{0.99} \\
 & Sampling, Few-shot & \cellcolor[RGB]{146,196,222}\textcolor{black}{0.28} & \cellcolor[RGB]{119,181,217}\textcolor{black}{0.72} & \cellcolor[RGB]{203,222,241}\textcolor{black}{0.95} & \cellcolor[RGB]{215,230,245}\textcolor{black}{0.99} & \cellcolor[RGB]{78,154,203}\textcolor{black}{0.97} & \cellcolor[RGB]{247,251,255}\textcolor{black}{0.99} & \cellcolor[RGB]{15,90,163}\textcolor{white}{0.99} & \cellcolor[RGB]{164,204,227}\textcolor{black}{0.98} \\
 & Persona & \cellcolor[RGB]{50,130,190}\textcolor{white}{0.24} & \cellcolor[RGB]{108,174,214}\textcolor{black}{0.73} & \cellcolor[RGB]{19,95,167}\textcolor{white}{0.98} & \cellcolor[RGB]{37,117,183}\textcolor{white}{1.00} & \cellcolor[RGB]{16,91,164}\textcolor{white}{0.99} & \cellcolor[RGB]{29,108,177}\textcolor{white}{1.00} & \cellcolor[RGB]{8,48,107}\textcolor{white}{1.00} & \cellcolor[RGB]{15,90,163}\textcolor{white}{0.99} \\
 & Persona, Few-shot & \cellcolor[RGB]{50,130,190}\textcolor{white}{0.24} & \cellcolor[RGB]{99,168,211}\textcolor{black}{0.73} & \cellcolor[RGB]{32,112,180}\textcolor{white}{0.97} & \cellcolor[RGB]{124,183,218}\textcolor{black}{0.99} & \cellcolor[RGB]{32,111,180}\textcolor{white}{0.98} & \cellcolor[RGB]{122,182,217}\textcolor{black}{0.99} & \cellcolor[RGB]{31,110,179}\textcolor{white}{0.98} & \cellcolor[RGB]{32,111,180}\textcolor{white}{0.99} \\
 & Batch & \cellcolor[RGB]{72,150,200}\textcolor{white}{0.25} & \cellcolor[RGB]{91,163,208}\textcolor{black}{0.74} & \cellcolor[RGB]{49,129,189}\textcolor{white}{0.97} & \cellcolor[RGB]{10,83,158}\textcolor{white}{1.00} & \cellcolor[RGB]{39,119,184}\textcolor{white}{0.98} & \cellcolor[RGB]{16,91,164}\textcolor{white}{1.00} & \cellcolor[RGB]{104,172,213}\textcolor{black}{0.96} & \cellcolor[RGB]{8,64,130}\textcolor{white}{1.00} \\
 & Batch, Few-shot & \cellcolor[RGB]{65,145,198}\textcolor{white}{0.25} & \cellcolor[RGB]{90,162,207}\textcolor{black}{0.74} & \cellcolor[RGB]{8,78,152}\textcolor{white}{0.98} & \cellcolor[RGB]{8,74,145}\textcolor{white}{1.00} & \cellcolor[RGB]{23,100,171}\textcolor{white}{0.98} & \cellcolor[RGB]{8,81,156}\textcolor{white}{1.00} & \cellcolor[RGB]{95,166,209}\textcolor{black}{0.96} & \cellcolor[RGB]{8,70,139}\textcolor{white}{1.00} \\
 & Description & \cellcolor[RGB]{8,76,149}\textcolor{white}{0.20} & \cellcolor[RGB]{101,170,212}\textcolor{black}{0.73} & \cellcolor[RGB]{130,187,219}\textcolor{black}{0.96} & \cellcolor[RGB]{31,110,179}\textcolor{white}{1.00} & \cellcolor[RGB]{102,171,212}\textcolor{black}{0.96} & \cellcolor[RGB]{33,113,181}\textcolor{white}{1.00} & \cellcolor[RGB]{171,208,230}\textcolor{black}{0.94} & \cellcolor[RGB]{11,85,159}\textcolor{white}{0.99} \\
 & Description, Few-shot & \cellcolor[RGB]{14,89,162}\textcolor{white}{0.21} & \cellcolor[RGB]{84,159,205}\textcolor{black}{0.75} & \cellcolor[RGB]{132,188,219}\textcolor{black}{0.96} & \cellcolor[RGB]{45,125,187}\textcolor{white}{1.00} & \cellcolor[RGB]{79,155,203}\textcolor{black}{0.97} & \cellcolor[RGB]{53,133,191}\textcolor{white}{1.00} & \cellcolor[RGB]{145,195,222}\textcolor{black}{0.95} & \cellcolor[RGB]{37,117,183}\textcolor{white}{0.99} \\
\midrule \multirow{8}{*}{Mistral} & Sampling & \cellcolor[RGB]{137,190,220}\textcolor{black}{0.28} & \cellcolor[RGB]{225,237,248}\textcolor{black}{0.61} & \cellcolor[RGB]{21,97,169}\textcolor{white}{0.98} & \cellcolor[RGB]{15,90,163}\textcolor{white}{1.00} & \cellcolor[RGB]{41,121,185}\textcolor{white}{0.98} & \cellcolor[RGB]{25,103,173}\textcolor{white}{1.00} & \cellcolor[RGB]{111,176,215}\textcolor{black}{0.96} & \cellcolor[RGB]{8,81,156}\textcolor{white}{1.00} \\
 & Sampling, Few-shot & \cellcolor[RGB]{127,185,218}\textcolor{black}{0.27} & \cellcolor[RGB]{227,238,248}\textcolor{black}{0.61} & \cellcolor[RGB]{52,132,191}\textcolor{white}{0.97} & \cellcolor[RGB]{14,89,162}\textcolor{white}{1.00} & \cellcolor[RGB]{55,135,192}\textcolor{white}{0.97} & \cellcolor[RGB]{25,103,173}\textcolor{white}{1.00} & \cellcolor[RGB]{173,208,230}\textcolor{black}{0.94} & \cellcolor[RGB]{8,68,136}\textcolor{white}{1.00} \\
 & Persona & \cellcolor[RGB]{36,116,183}\textcolor{white}{0.23} & \cellcolor[RGB]{193,217,237}\textcolor{black}{0.66} & \cellcolor[RGB]{46,126,188}\textcolor{white}{0.97} & \cellcolor[RGB]{12,86,160}\textcolor{white}{1.00} & \cellcolor[RGB]{38,118,184}\textcolor{white}{0.98} & \cellcolor[RGB]{18,94,166}\textcolor{white}{1.00} & \cellcolor[RGB]{233,242,250}\textcolor{black}{0.92} & \cellcolor[RGB]{30,109,178}\textcolor{white}{0.99} \\
 & Persona, Few-shot & \cellcolor[RGB]{54,134,192}\textcolor{white}{0.24} & \cellcolor[RGB]{203,222,241}\textcolor{black}{0.65} & \cellcolor[RGB]{58,138,194}\textcolor{white}{0.97} & \cellcolor[RGB]{8,77,150}\textcolor{white}{1.00} & \cellcolor[RGB]{45,125,187}\textcolor{white}{0.98} & \cellcolor[RGB]{18,93,166}\textcolor{white}{1.00} & \cellcolor[RGB]{180,211,233}\textcolor{black}{0.94} & \cellcolor[RGB]{8,59,124}\textcolor{white}{1.00} \\
 & Batch & \cellcolor[RGB]{239,246,252}\textcolor{black}{0.34} & \cellcolor[RGB]{236,244,251}\textcolor{black}{0.60} & \cellcolor[RGB]{127,185,218}\textcolor{black}{0.96} & \cellcolor[RGB]{209,226,243}\textcolor{black}{0.99} & \cellcolor[RGB]{95,166,209}\textcolor{black}{0.97} & \cellcolor[RGB]{183,212,234}\textcolor{black}{0.99} & \cellcolor[RGB]{170,207,229}\textcolor{black}{0.94} & \cellcolor[RGB]{65,145,198}\textcolor{white}{0.99} \\
 & Batch, Few-shot & \cellcolor[RGB]{247,251,255}\textcolor{black}{0.35} & \cellcolor[RGB]{225,237,248}\textcolor{black}{0.61} & \cellcolor[RGB]{198,219,239}\textcolor{black}{0.95} & \cellcolor[RGB]{224,236,248}\textcolor{black}{0.99} & \cellcolor[RGB]{125,184,218}\textcolor{black}{0.96} & \cellcolor[RGB]{140,192,221}\textcolor{black}{0.99} & \cellcolor[RGB]{225,237,248}\textcolor{black}{0.92} & \cellcolor[RGB]{125,184,218}\textcolor{black}{0.99} \\
 & Description & \cellcolor[RGB]{8,61,127}\textcolor{white}{0.19} & \cellcolor[RGB]{90,162,207}\textcolor{black}{0.74} & \cellcolor[RGB]{52,132,191}\textcolor{white}{0.97} & \cellcolor[RGB]{31,110,179}\textcolor{white}{1.00} & \cellcolor[RGB]{46,126,188}\textcolor{white}{0.98} & \cellcolor[RGB]{36,116,183}\textcolor{white}{1.00} & \cellcolor[RGB]{101,170,212}\textcolor{black}{0.96} & \cellcolor[RGB]{25,102,173}\textcolor{white}{0.99} \\
 & Description, Few-shot & \cellcolor[RGB]{48,128,189}\textcolor{white}{0.24} & \cellcolor[RGB]{212,228,244}\textcolor{black}{0.63} & \cellcolor[RGB]{220,234,246}\textcolor{black}{0.94} & \cellcolor[RGB]{217,231,245}\textcolor{black}{0.99} & \cellcolor[RGB]{169,207,229}\textcolor{black}{0.95} & \cellcolor[RGB]{208,225,242}\textcolor{black}{0.99} & \cellcolor[RGB]{216,231,245}\textcolor{black}{0.93} & \cellcolor[RGB]{102,171,212}\textcolor{black}{0.99} \\
\midrule \multirow{5}{*}{Baseline} & Random & \cellcolor[RGB]{41,121,185}\textcolor{white}{0.23} & \cellcolor[RGB]{247,251,255}\textcolor{black}{0.58} & \cellcolor[RGB]{247,251,255}\textcolor{black}{0.94} & \cellcolor[RGB]{247,251,255}\textcolor{black}{0.99} & \cellcolor[RGB]{247,251,255}\textcolor{black}{0.92} & \cellcolor[RGB]{208,226,242}\textcolor{black}{0.99} & \cellcolor[RGB]{247,251,255}\textcolor{black}{0.91} & \cellcolor[RGB]{186,214,235}\textcolor{black}{0.98} \\
 & Popularity score &  & \cellcolor[RGB]{96,167,210}\textcolor{black}{0.74} & \cellcolor[RGB]{21,97,169}\textcolor{white}{0.98} & \cellcolor[RGB]{8,48,107}\textcolor{white}{1.00} & \cellcolor[RGB]{8,48,107}\textcolor{white}{1.00} & \cellcolor[RGB]{8,48,107}\textcolor{white}{1.00} & \cellcolor[RGB]{183,212,234}\textcolor{black}{0.94} & \cellcolor[RGB]{247,251,255}\textcolor{black}{0.98} \\
 & 5 real data & \cellcolor[RGB]{166,206,228}\textcolor{black}{0.29} & \cellcolor[RGB]{127,185,218}\textcolor{black}{0.71} & \cellcolor[RGB]{70,149,200}\textcolor{white}{0.97} & \cellcolor[RGB]{185,214,234}\textcolor{black}{0.99} & \cellcolor[RGB]{83,158,205}\textcolor{black}{0.97} & \cellcolor[RGB]{114,178,216}\textcolor{black}{0.99} & \cellcolor[RGB]{106,174,214}\textcolor{black}{0.96} & \cellcolor[RGB]{39,119,184}\textcolor{white}{0.99} \\
 & 15 real data & \cellcolor[RGB]{51,131,190}\textcolor{white}{0.24} & \cellcolor[RGB]{53,133,191}\textcolor{white}{0.78} & \cellcolor[RGB]{15,90,163}\textcolor{white}{0.98} & \cellcolor[RGB]{37,117,183}\textcolor{white}{1.00} & \cellcolor[RGB]{22,99,170}\textcolor{white}{0.99} & \cellcolor[RGB]{58,138,194}\textcolor{white}{1.00} & \cellcolor[RGB]{8,62,129}\textcolor{white}{1.00} & \cellcolor[RGB]{8,64,130}\textcolor{white}{1.00} \\
 & 30 real data & \cellcolor[RGB]{8,51,112}\textcolor{white}{0.19} & \cellcolor[RGB]{8,48,107}\textcolor{white}{0.87} & \cellcolor[RGB]{8,48,107}\textcolor{white}{0.99} & \cellcolor[RGB]{21,98,169}\textcolor{white}{1.00} & \cellcolor[RGB]{8,79,153}\textcolor{white}{0.99} & \cellcolor[RGB]{16,91,164}\textcolor{white}{1.00} & \cellcolor[RGB]{8,74,145}\textcolor{white}{0.99} & \cellcolor[RGB]{8,48,107}\textcolor{white}{1.00} \\
\bottomrule
\end{tabular}
\caption{Assortment optimization results for position-dependent rewards $r_p = 10-0.1 (p-1)^2$, averaged over 20 runs. Entries report the mean value for each model and method. Cell shading is column-wise, with darker indicating better performance within that metric. The model ``Gemini'' refers to Gemini 3 Flash, and ``Mistral'' refers to Mistral Large 3.}
\label{tab:assortment_results_10-0.1*i*i}
\end{table*}

\clearpage
\section{Pricing problem: Generation Details and Full Results} \label{app:pricingTables}

We consider 20 different distributions $\hF$ for each LLM generation method and each baseline, calling them multiple times to handle the randomness in the distribution generated.
\begin{itemize}
\item \textbf{Sampling}: We generate a pool of 100 willingness-to-pay values for each product using the LLM. We subsample 50 willingness-to-pay values to form an estimated distribution $\hF$, and repeat to form 20 estimated distributions in this way.
\item \textbf{Persona-sampling}:  We generate a pool of 100 willingness-to-pay values for each products, prompting the LLM with a different persona each time. We subsample 50 willingness-to-pay values to form an estimated distribution $\hF$, and repeat to form 20 estimated distributions in this way.
\item \textbf{Batch-generation}:  We generate 25 willingness-to-pay values for each product per query and repeat this process 20 times, yielding 20 distributions.
\item \textbf{Description}: We ask the LLM to generate 5 most likely willingness-to-pay values and their corresponding probabilities for each product, repeating 20 times. 
\item \textbf{Few-shot Sampling}: We draw 5 example sets, each containing 6 examples. For each example set, we generate a pool of 100 willingness-to-pay values for each product and construct 4 subsamples of size 50, resulting in a total of 20 distributions.
\item \textbf{Few-shot Persona-sampling}: Same as Few-shot Sampling, except we also prompt the LLM with a different persona each time.
\item \textbf{Few-shot Batch-generation}: We draw 5 example sets, each containing 6 examples.  We run Batch-generation 4 times with each example set, resulting in a total of 20 distributions.
\item \textbf{Few-shot Description}: We draw 5 example sets, each containing 6 examples. We run Description 4 times with each example set, resulting in a total of 20 distributions.
\item \textbf{Random baseline}: We generate a pool of 100 willingness-to-pay values for each product uniformly at random.  We subsample 50 willingness-to-pay values from it 20 times, to form 20 estimated distributions.
\item \textbf{$d$ real data baseline}: We randomly sample $d$ willingness-to-pay values from the ground-truth $F$ for each product, generating 20 distributions in this way.
\end{itemize}

The reported values take an average over the 6 products, with different ground-truth distributions $F$.

\begin{table*}[h]
\footnotesize
\centering
\begin{tabular}{l l c c c c c c}
\toprule
 && \multirow{1}{*}{\textbf{Wasserstein}} & \multirow{1}{*}{\textbf{Kolmogorov}}& \textbf{WorstCR} & \multicolumn{3}{c}{\textbf{AverageCR} \textit{(higher is better)}} \\
 \cmidrule(lr){5-5} \cmidrule(lr){6-8}
 && \textit{(lower is} & \textit{(lower is}  & \textit{(higher is} & \multirow{2}{*}{$\mathrm{Unif[0, 32]}$} & \multirow{2}{*}{$\mathrm{Unif[0, 66]}$} & \multirow{2}{*}{$\mathrm{Unif[0, 100]}$} \\
\textbf{Model} & \textbf{Method} & \textit{better)} &  \textit{better)} &  \textit{better)} & & & \\
 & & & & \scriptsize $c \in [0, 32]$ & & & \\
\midrule \multirow{8}{*}{GPT-4o} & Sampling & \cellcolor[RGB]{193,217,237}\textcolor{black}{17.13} & \cellcolor[RGB]{170,207,229}\textcolor{black}{0.42} & \cellcolor[RGB]{8,48,107}\textcolor{white}{0.85} & \cellcolor[RGB]{12,86,160}\textcolor{white}{0.92} & \cellcolor[RGB]{143,194,222}\textcolor{black}{0.78} & \cellcolor[RGB]{153,199,224}\textcolor{black}{0.61} \\
 & Sampling, Few-shot & \cellcolor[RGB]{179,211,232}\textcolor{black}{16.51} & \cellcolor[RGB]{154,200,224}\textcolor{black}{0.41} & \cellcolor[RGB]{8,58,122}\textcolor{white}{0.84} & \cellcolor[RGB]{8,70,139}\textcolor{white}{0.94} & \cellcolor[RGB]{157,202,225}\textcolor{black}{0.77} & \cellcolor[RGB]{166,206,228}\textcolor{black}{0.60} \\
 & Sampling, Persona & \cellcolor[RGB]{161,203,226}\textcolor{black}{15.66} & \cellcolor[RGB]{156,201,225}\textcolor{black}{0.41} & \cellcolor[RGB]{8,59,124}\textcolor{white}{0.84} & \cellcolor[RGB]{8,48,107}\textcolor{white}{0.96} & \cellcolor[RGB]{116,179,216}\textcolor{black}{0.79} & \cellcolor[RGB]{116,179,216}\textcolor{black}{0.65} \\
 & Persona, Few-shot & \cellcolor[RGB]{200,220,240}\textcolor{black}{17.57} & \cellcolor[RGB]{176,210,231}\textcolor{black}{0.43} & \cellcolor[RGB]{32,112,180}\textcolor{white}{0.78} & \cellcolor[RGB]{8,65,132}\textcolor{white}{0.94} & \cellcolor[RGB]{178,210,232}\textcolor{black}{0.75} & \cellcolor[RGB]{143,194,222}\textcolor{black}{0.62} \\
 & Batch & \cellcolor[RGB]{113,177,215}\textcolor{black}{13.84} & \cellcolor[RGB]{73,151,201}\textcolor{white}{0.33} & \cellcolor[RGB]{125,184,218}\textcolor{black}{0.70} & \cellcolor[RGB]{44,124,186}\textcolor{white}{0.89} & \cellcolor[RGB]{127,185,218}\textcolor{black}{0.79} & \cellcolor[RGB]{97,167,210}\textcolor{black}{0.67} \\
 & Batch, Few-shot & \cellcolor[RGB]{100,169,211}\textcolor{black}{13.33} & \cellcolor[RGB]{61,141,196}\textcolor{white}{0.31} & \cellcolor[RGB]{102,171,212}\textcolor{black}{0.72} & \cellcolor[RGB]{43,123,186}\textcolor{white}{0.89} & \cellcolor[RGB]{97,167,210}\textcolor{black}{0.80} & \cellcolor[RGB]{77,153,202}\textcolor{black}{0.69} \\
 & Description & \cellcolor[RGB]{74,152,201}\textcolor{white}{12.13} & \cellcolor[RGB]{60,140,195}\textcolor{white}{0.31} & \cellcolor[RGB]{59,139,194}\textcolor{white}{0.75} & \cellcolor[RGB]{55,135,192}\textcolor{white}{0.88} & \cellcolor[RGB]{68,147,199}\textcolor{white}{0.83} & \cellcolor[RGB]{34,114,182}\textcolor{white}{0.75} \\
 & Description, Few-shot & \cellcolor[RGB]{79,155,203}\textcolor{black}{12.40} & \cellcolor[RGB]{63,143,197}\textcolor{white}{0.31} & \cellcolor[RGB]{44,124,186}\textcolor{white}{0.77} & \cellcolor[RGB]{22,99,170}\textcolor{white}{0.91} & \cellcolor[RGB]{72,150,200}\textcolor{white}{0.82} & \cellcolor[RGB]{51,131,190}\textcolor{white}{0.73} \\
\midrule \multirow{8}{*}{GPT-5-mini} & Sampling & \cellcolor[RGB]{213,229,244}\textcolor{black}{18.57} & \cellcolor[RGB]{202,222,240}\textcolor{black}{0.46} & \cellcolor[RGB]{8,75,147}\textcolor{white}{0.82} & \cellcolor[RGB]{8,67,135}\textcolor{white}{0.94} & \cellcolor[RGB]{201,221,240}\textcolor{black}{0.74} & \cellcolor[RGB]{193,217,237}\textcolor{black}{0.56} \\
 & Sampling, Few-shot & \cellcolor[RGB]{199,219,239}\textcolor{black}{17.43} & \cellcolor[RGB]{181,212,233}\textcolor{black}{0.44} & \cellcolor[RGB]{10,83,158}\textcolor{white}{0.81} & \cellcolor[RGB]{31,110,179}\textcolor{white}{0.90} & \cellcolor[RGB]{140,192,221}\textcolor{black}{0.78} & \cellcolor[RGB]{175,209,231}\textcolor{black}{0.59} \\
 & Sampling, Persona & \cellcolor[RGB]{193,217,237}\textcolor{black}{17.13} & \cellcolor[RGB]{174,209,231}\textcolor{black}{0.43} & \cellcolor[RGB]{8,72,142}\textcolor{white}{0.83} & \cellcolor[RGB]{8,60,125}\textcolor{white}{0.95} & \cellcolor[RGB]{154,200,224}\textcolor{black}{0.77} & \cellcolor[RGB]{170,207,229}\textcolor{black}{0.59} \\
 & Persona, Few-shot & \cellcolor[RGB]{69,148,199}\textcolor{white}{11.90} & \cellcolor[RGB]{61,141,196}\textcolor{white}{0.31} & \cellcolor[RGB]{14,89,162}\textcolor{white}{0.81} & \cellcolor[RGB]{36,116,183}\textcolor{white}{0.90} & \cellcolor[RGB]{72,150,200}\textcolor{white}{0.82} & \cellcolor[RGB]{53,133,191}\textcolor{white}{0.72} \\
 & Batch & \cellcolor[RGB]{60,140,195}\textcolor{white}{11.48} & \cellcolor[RGB]{44,124,186}\textcolor{white}{0.29} & \cellcolor[RGB]{151,198,223}\textcolor{black}{0.68} & \cellcolor[RGB]{52,132,191}\textcolor{white}{0.88} & \cellcolor[RGB]{63,143,197}\textcolor{white}{0.83} & \cellcolor[RGB]{43,123,186}\textcolor{white}{0.74} \\
 & Batch, Few-shot & \cellcolor[RGB]{44,124,186}\textcolor{white}{10.54} & \cellcolor[RGB]{23,100,171}\textcolor{white}{0.25} & \cellcolor[RGB]{108,174,214}\textcolor{black}{0.71} & \cellcolor[RGB]{43,123,186}\textcolor{white}{0.89} & \cellcolor[RGB]{66,146,198}\textcolor{white}{0.83} & \cellcolor[RGB]{32,111,180}\textcolor{white}{0.76} \\
 & Description & \cellcolor[RGB]{96,167,210}\textcolor{black}{13.15} & \cellcolor[RGB]{81,156,204}\textcolor{black}{0.33} & \cellcolor[RGB]{42,122,185}\textcolor{white}{0.77} & \cellcolor[RGB]{18,94,166}\textcolor{white}{0.92} & \cellcolor[RGB]{49,129,189}\textcolor{white}{0.84} & \cellcolor[RGB]{33,113,181}\textcolor{white}{0.76} \\
 & Description, Few-shot & \cellcolor[RGB]{91,163,208}\textcolor{black}{12.94} & \cellcolor[RGB]{81,156,204}\textcolor{black}{0.33} & \cellcolor[RGB]{50,130,190}\textcolor{white}{0.76} & \cellcolor[RGB]{8,80,155}\textcolor{white}{0.93} & \cellcolor[RGB]{35,115,182}\textcolor{white}{0.85} & \cellcolor[RGB]{21,98,169}\textcolor{white}{0.78} \\
\midrule \multirow{8}{*}{Gemini} & Sampling & \cellcolor[RGB]{174,209,231}\textcolor{black}{16.25} & \cellcolor[RGB]{201,221,240}\textcolor{black}{0.46} & \cellcolor[RGB]{22,99,170}\textcolor{white}{0.80} & \cellcolor[RGB]{47,127,188}\textcolor{white}{0.89} & \cellcolor[RGB]{113,177,215}\textcolor{black}{0.79} & \cellcolor[RGB]{97,167,210}\textcolor{black}{0.67} \\
 & Sampling, Few-shot & \cellcolor[RGB]{151,198,223}\textcolor{black}{15.28} & \cellcolor[RGB]{148,196,223}\textcolor{black}{0.40} & \cellcolor[RGB]{34,114,182}\textcolor{white}{0.78} & \cellcolor[RGB]{55,135,192}\textcolor{white}{0.88} & \cellcolor[RGB]{36,116,183}\textcolor{white}{0.85} & \cellcolor[RGB]{15,90,163}\textcolor{white}{0.79} \\
 & Sampling, Persona & \cellcolor[RGB]{121,181,217}\textcolor{black}{14.17} & \cellcolor[RGB]{100,169,211}\textcolor{black}{0.36} & \cellcolor[RGB]{8,60,125}\textcolor{white}{0.84} & \cellcolor[RGB]{24,101,172}\textcolor{white}{0.91} & \cellcolor[RGB]{127,185,218}\textcolor{black}{0.79} & \cellcolor[RGB]{121,181,217}\textcolor{black}{0.64} \\
 & Persona, Few-shot & \cellcolor[RGB]{199,219,239}\textcolor{black}{17.44} & \cellcolor[RGB]{179,211,232}\textcolor{black}{0.43} & \cellcolor[RGB]{43,123,186}\textcolor{white}{0.77} & \cellcolor[RGB]{35,115,182}\textcolor{white}{0.90} & \cellcolor[RGB]{211,227,243}\textcolor{black}{0.72} & \cellcolor[RGB]{196,218,238}\textcolor{black}{0.56} \\
 & Batch & \cellcolor[RGB]{25,102,173}\textcolor{white}{9.31} & \cellcolor[RGB]{12,86,160}\textcolor{white}{0.23} & \cellcolor[RGB]{96,167,210}\textcolor{black}{0.72} & \cellcolor[RGB]{38,118,184}\textcolor{white}{0.89} & \cellcolor[RGB]{82,157,204}\textcolor{black}{0.82} & \cellcolor[RGB]{48,128,189}\textcolor{white}{0.73} \\
 & Batch, Few-shot & \cellcolor[RGB]{39,119,184}\textcolor{white}{10.29} & \cellcolor[RGB]{24,101,172}\textcolor{white}{0.26} & \cellcolor[RGB]{93,165,209}\textcolor{black}{0.73} & \cellcolor[RGB]{35,115,182}\textcolor{white}{0.90} & \cellcolor[RGB]{62,142,196}\textcolor{white}{0.83} & \cellcolor[RGB]{33,113,181}\textcolor{white}{0.75} \\
 & Description & \cellcolor[RGB]{119,181,217}\textcolor{black}{14.12} & \cellcolor[RGB]{129,186,219}\textcolor{black}{0.38} & \cellcolor[RGB]{19,95,167}\textcolor{white}{0.80} & \cellcolor[RGB]{17,92,165}\textcolor{white}{0.92} & \cellcolor[RGB]{105,173,213}\textcolor{black}{0.80} & \cellcolor[RGB]{96,167,210}\textcolor{black}{0.67} \\
 & Description, Few-shot & \cellcolor[RGB]{166,206,228}\textcolor{black}{15.90} & \cellcolor[RGB]{170,207,229}\textcolor{black}{0.42} & \cellcolor[RGB]{25,102,173}\textcolor{white}{0.79} & \cellcolor[RGB]{10,84,158}\textcolor{white}{0.93} & \cellcolor[RGB]{181,212,233}\textcolor{black}{0.75} & \cellcolor[RGB]{179,211,232}\textcolor{black}{0.58} \\
\midrule \multirow{8}{*}{Mistral} & Sampling & \cellcolor[RGB]{194,217,238}\textcolor{black}{17.23} & \cellcolor[RGB]{176,210,231}\textcolor{black}{0.43} & \cellcolor[RGB]{8,78,152}\textcolor{white}{0.82} & \cellcolor[RGB]{8,55,118}\textcolor{white}{0.95} & \cellcolor[RGB]{163,204,227}\textcolor{black}{0.76} & \cellcolor[RGB]{165,205,227}\textcolor{black}{0.60} \\
 & Sampling, Few-shot & \cellcolor[RGB]{194,217,238}\textcolor{black}{17.19} & \cellcolor[RGB]{189,215,236}\textcolor{black}{0.44} & \cellcolor[RGB]{14,88,162}\textcolor{white}{0.81} & \cellcolor[RGB]{9,82,157}\textcolor{white}{0.93} & \cellcolor[RGB]{151,198,223}\textcolor{black}{0.77} & \cellcolor[RGB]{200,220,240}\textcolor{black}{0.55} \\
 & Sampling, Persona & \cellcolor[RGB]{247,251,255}\textcolor{black}{21.17} & \cellcolor[RGB]{247,251,255}\textcolor{black}{0.55} & \cellcolor[RGB]{81,156,204}\textcolor{black}{0.74} & \cellcolor[RGB]{10,83,158}\textcolor{white}{0.93} & \cellcolor[RGB]{247,251,255}\textcolor{black}{0.68} & \cellcolor[RGB]{247,251,255}\textcolor{black}{0.46} \\
 & Persona, Few-shot & \cellcolor[RGB]{218,232,246}\textcolor{black}{18.95} & \cellcolor[RGB]{207,225,242}\textcolor{black}{0.47} & \cellcolor[RGB]{54,134,192}\textcolor{white}{0.76} & \cellcolor[RGB]{8,70,139}\textcolor{white}{0.94} & \cellcolor[RGB]{208,226,242}\textcolor{black}{0.73} & \cellcolor[RGB]{221,234,247}\textcolor{black}{0.51} \\
 & Batch & \cellcolor[RGB]{195,218,238}\textcolor{black}{17.27} & \cellcolor[RGB]{138,191,221}\textcolor{black}{0.39} & \cellcolor[RGB]{205,224,241}\textcolor{black}{0.63} & \cellcolor[RGB]{64,144,197}\textcolor{white}{0.87} & \cellcolor[RGB]{173,208,230}\textcolor{black}{0.76} & \cellcolor[RGB]{124,183,218}\textcolor{black}{0.64} \\
 & Batch, Few-shot & \cellcolor[RGB]{171,208,230}\textcolor{black}{16.12} & \cellcolor[RGB]{104,172,213}\textcolor{black}{0.36} & \cellcolor[RGB]{190,216,236}\textcolor{black}{0.65} & \cellcolor[RGB]{54,134,192}\textcolor{white}{0.88} & \cellcolor[RGB]{179,211,232}\textcolor{black}{0.75} & \cellcolor[RGB]{148,196,223}\textcolor{black}{0.62} \\
 & Description & \cellcolor[RGB]{141,193,221}\textcolor{black}{14.93} & \cellcolor[RGB]{121,181,217}\textcolor{black}{0.38} & \cellcolor[RGB]{36,116,183}\textcolor{white}{0.78} & \cellcolor[RGB]{16,91,164}\textcolor{white}{0.92} & \cellcolor[RGB]{127,185,218}\textcolor{black}{0.79} & \cellcolor[RGB]{106,174,214}\textcolor{black}{0.66} \\
 & Description, Few-shot & \cellcolor[RGB]{201,221,240}\textcolor{black}{17.65} & \cellcolor[RGB]{168,206,228}\textcolor{black}{0.42} & \cellcolor[RGB]{72,150,200}\textcolor{white}{0.74} & \cellcolor[RGB]{10,83,158}\textcolor{white}{0.93} & \cellcolor[RGB]{198,219,239}\textcolor{black}{0.74} & \cellcolor[RGB]{181,212,233}\textcolor{black}{0.58} \\
\midrule \multirow{4}{*}{Baseline} & Random & \cellcolor[RGB]{244,249,254}\textcolor{black}{20.89} & \cellcolor[RGB]{99,168,211}\textcolor{black}{0.35} & \cellcolor[RGB]{247,251,255}\textcolor{black}{0.58} & \cellcolor[RGB]{247,251,255}\textcolor{black}{0.72} & \cellcolor[RGB]{247,251,255}\textcolor{black}{0.68} & \cellcolor[RGB]{97,167,210}\textcolor{black}{0.67} \\
 & 5 real data & \cellcolor[RGB]{69,148,199}\textcolor{white}{11.93} & \cellcolor[RGB]{84,159,205}\textcolor{black}{0.34} & \cellcolor[RGB]{78,154,203}\textcolor{black}{0.74} & \cellcolor[RGB]{58,138,194}\textcolor{white}{0.88} & \cellcolor[RGB]{96,167,210}\textcolor{black}{0.80} & \cellcolor[RGB]{73,151,201}\textcolor{white}{0.70} \\
 & 10 real data & \cellcolor[RGB]{14,88,162}\textcolor{white}{8.49} & \cellcolor[RGB]{18,93,166}\textcolor{white}{0.24} & \cellcolor[RGB]{28,106,176}\textcolor{white}{0.79} & \cellcolor[RGB]{32,111,180}\textcolor{white}{0.90} & \cellcolor[RGB]{18,94,166}\textcolor{white}{0.87} & \cellcolor[RGB]{14,88,162}\textcolor{white}{0.79} \\
 & 20 real data & \cellcolor[RGB]{8,48,107}\textcolor{white}{6.15} & \cellcolor[RGB]{8,48,107}\textcolor{white}{0.18} & \cellcolor[RGB]{8,62,129}\textcolor{white}{0.84} & \cellcolor[RGB]{8,73,144}\textcolor{white}{0.94} & \cellcolor[RGB]{8,48,107}\textcolor{white}{0.91} & \cellcolor[RGB]{8,48,107}\textcolor{white}{0.86} \\
\bottomrule
\end{tabular}
\caption{Pricing results over 6 products, each product averaged over 20 runs. Entries report the mean value over the 6 products for each model and method. Cell shading is column-wise, with darker indicating better performance within that metric. The model ``Gemini'' refers to Gemini 3 Flash, and ``Mistral'' refers to Mistral Large 3.}
\label{tab:pricing_results_all}
\end{table*}

\clearpage
\section{Plots of Willigness-to-Pay Distributions} 
\label{app:wtpPlots}

Recall that $R_c(a)=(a-c)\Pr_{\xi\sim F}[\xi\ge a]$ in the pricing problem.  We will refer to $\Pr_{\xi\sim F}[\xi\ge a]$ as the \textit{survival function} of a willingness-to-pay distribution $F$, depicting the probability of purchase as a function of the price $a$.

\begin{figure}[H]
    \centering
    \includegraphics[width=\linewidth]{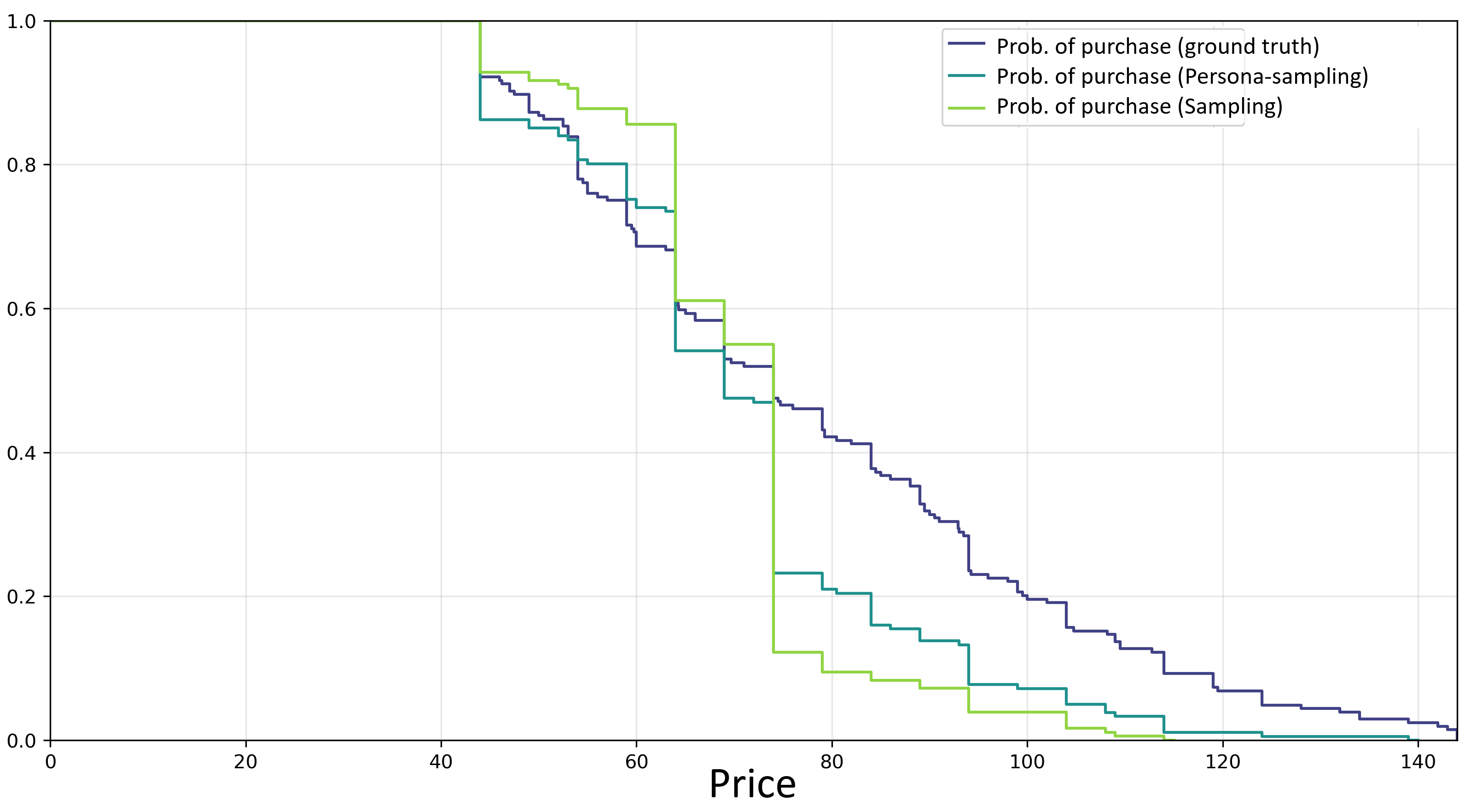}
    \caption{
    Survival functions of the willingness-to-pay distributions for the ground truth distribution $F$, and the estimated distributions $\hat F$ from the Sampling and Persona sampling methods for the product \emph{Bohol}.
    The ``drops'' in the functions are at technically at multiples of 4 and 9 --- this is because in the data, respondents were asked about their willingness-to-pay \textit{premiums} over the base price of PhP 44 (for an ``upgraded'' product), and this figure plots their total willingness-to-pay (44 + their response). 
}
    \label{fig:pricing_distribution}
\end{figure}

\Cref{fig:pricing_distribution} plots an example of the survival functions of the ground truth willingness-to-pay distribution $F$ and the estimated distributions $\hat{F}$ for the Sampling and Persona sampling methods.
A notable feature is that the true and estimated curves exhibit sharp ``drops'' at many of the same $x$-values.
These drops occur at willingness-to-pay values that are multiples of 5, 
reflecting the common phenomenon that when asked for willingness to pay, people disproportionately report round numbers ending in 0 or 5 ---  and we see that the LLM-generated samples display the same phenomenon.

This matters for pricing because the optimal price often lies exactly at a drop in the survival function.
A drop in the survival function at price $p$ corresponds to a drop in the demand when the price increases past $p$, and hence the optimal price will often occur right at the drop.
Consequently, when $\hat F$ places drops at the same prices as $F$, the maximizing price is more likely to coincide with the optimal price under $F$.
This provides intuition for how the LLM's world knowledge about round-number reporting can translate into improved downstream pricing decisions.

\clearpage
\section{Newsvendor problem: Data Pre-processing} \label{app:invPreprocessing}

We use a sales dataset at the fashion retailer H\&M  \citep{Kaggle_HM_Recommendations_2022}, which contains daily sales numbers at the product level.
We first aggregate the sales into weekly totals, and we focus on items categorized as \textbf{Trousers} over a 30-week period, from March 3, 2019, to September 22, 2019. 
Each item is associated with metadata providing a description of the product, which includes the fields: product name, product type name, graphical appearance name, color group name, department name, index name, section name, garment group name, and detailed description.

We used the following data pre-processing pipeline to ensure we only consider items without large price fluctuations or missing values:
\begin{itemize}
    \item \textbf{Price Outlier Removal:} To mitigate the impact of pricing anomalies or data entry errors, any price $P_{i}$ falling outside a 20\% threshold of the mean price $\mu_{p}$ was flagged as a missing value (NA). Specifically, values were discarded if:
    \[ |P_{i} - \mu_{p}| > 0.20 \mu_{p}. \]
    
    \item \textbf{Zero-Value Removal:} Rows containing a value of $0$ in any weekly sales count column were removed. This process excludes products that were likely inactive or out of stock during the observed period.
    
    \item \textbf{Missing Value Threshold:} We enforced a data density requirement by calculating the total number of NA entries per product record. Any item missing more than 10 weekly data points ($\text{NAs} > 10$) was excluded.
\end{itemize}

Following the application of these filters, the final refined dataset consists of \textbf{300 items}.

\clearpage
\section{Assortment problem: Prompts} \label{app:promptAsst}


\begin{enumerate}
    \item \textbf{Background Information:}
    \begin{quote}
    ``We study sushi preference patterns among Japanese consumers. The sushi items listed below are those considered. 
    
    \textbf{Sushi items (Target items):} 
    \begin{itemize}
        \item \texttt{0}: \{sushi 0 description\}
        \item \texttt{1}: \{sushi 1 description\}
        \item ...
        \item \texttt{9}: \{sushi 9 description\}
    \end{itemize}
    \textbf{Sushi item descriptions:}
    {\footnotesize
\begin{itemize}
  \setlength{\itemsep}{0pt}
  \setlength{\parskip}{0pt}
  \setlength{\parsep}{0pt}
  \item \texttt{Name and ID}
  \item \texttt{Style:} maki roll or non-maki
  \item \texttt{Major category:} seafood or non-seafood
  \item \texttt{Minor group:} one of 12 ingredient groups
  \item \texttt{Taste intensity:} very heavy / heavy / moderate / light / very light
  \item \texttt{Consumption frequency:} rarely / sometimes / often / very frequently eaten
  \item \texttt{Availability:} very rarely to very commonly found
  \item \texttt{Price score:} from 0 to 6
\end{itemize}
}

    \end{quote}

    \item \textbf{Additional Information in Few-shot Setting:}
    \begin{quote}
    ``To help with the estimation task, we provide some examples of a respondent’s decision as guidance for the model. 
    
    The examples are listed below:

    \textbf{Examples (totally 6 examples):} 
    \begin{itemize}
        \item \texttt{0}: A person made a decision of: \{a ranking\}
        \item \texttt{1}: A person made a decision of: \{a ranking\}
        \item ...
        \item \texttt{5}: A person made a decision of: \{a ranking\}
    \end{itemize}

    \end{quote}

    
    

    
    
\end{enumerate}

For each of the two information settings, we test four ways of prompting an LLM to generate a distribution.

\begin{enumerate}
    \item \textbf{Sampling:}
    \begin{quote}
        ``Based on the information provided, please simulate a Japanese respondent’s sushi preference ranking. 
        
        \textbf{Task:} Generate one ranking to represent a potential preference ordering over the 10 sushi items.
        
        \textbf{INSTRUCTIONS:}
        \begin{enumerate}
            \item First, explain your reasoning. Consider the sushi item descriptions (and examples and persona if we provided).
            \item Second, provide the final ranking as exactly 10 unique integers from 0 to 9, ordered from most preferred to least preferred.
        \end{enumerate}
        \textbf{Output Format:} Reasoning: [...]
Final Answer: [10 integers separated by spaces]

    \end{quote}

    \item \textbf{Persona-sampling:} Same as sampling, but we provide the following additional instructions.
    \begin{quote}
    ``To help with the estimation task, we provide a persona description of the one you should pretend to be. 

    So, pretend you are \{persona description\}.

    \textbf{Persona description (outside the distribution): } 
    {\footnotesize
\begin{itemize}
  \setlength{\itemsep}{0pt}
  \setlength{\parskip}{0pt}
  \setlength{\parsep}{0pt}
  \item \texttt {Gender and age group}, where age is discretized into six bins (15--19, 20--29, 30--39, 40--49, 50--59, 60+)
  \item \texttt {Current residence}, including prefecture, region, and east--west classification within Japan.
  \item \texttt {Childhood residence}, including prefecture, region, and east--west classification, when available.
\end{itemize}
}

    \end{quote}

    \item \textbf{Batch-generation:}
    \begin{quote}
    ``Based on the information provided, please generate a representative batch of sushi preference rankings.

    \textbf{Task:} Generate a batch of 30 independent preference rankings over the 10 sushi items.

     \textbf{INSTRUCTIONS:}
        \begin{enumerate}
            \item First, explain your reasoning. Consider the sushi item descriptions (and examples if we provided).
            \item Second, please output exactly 30 rankings, each consisting of 10 unique integers from 0 to 9, ordered from most preferred to least preferred.
        \end{enumerate}
        \textbf{Output Format:} Reasoning: [...]
Final Answer: [30 lines of rankings, each line corresponding to one ranking.]

    \end{quote}

    \item \textbf{Description:}
    \begin{quote}
        ``Based on the information provided, please verbalize the utilities for the 10 sushi.

    \textbf{Task:} Provide utilities to 10 sushi, and utilities can be any real numbers (scale/shift invariant) and must be floats in [0, 5].

     \textbf{INSTRUCTIONS:}
        \begin{enumerate}
            \item First, explain your reasoning. Consider the sushi item descriptions (and examples if we provided).
            \item Second, provide the utilities.
        \end{enumerate}
        \textbf{Output Format:} Reasoning: [...]
Final Answer: [{\footnotesize
\texttt{
"0": <float>, "1": <float>, ..., "9": <float>
}}
]
    \end{quote}
    
\end{enumerate}

\clearpage
\section{Pricing problem: Prompts} \label{app:promptPricing}

\begin{enumerate}
    \item \textbf{Background Information:}
    \begin{quote}

    ``We study willingness-to-pay premiums for tablea chocolate products among consumers from Central Bicol State University of Agriculture in the Philippines. You will help us simulate willingness-to-pay decision. Here are three target items with there awards/origin:

    \textbf{Chocolate products (providing awards):} 
    \begin{itemize}
        \item \texttt{Bohol}: won Academy of Chocolate
        \item \texttt{Davao}: won Great Taste
        \item \texttt{Improved Bicol}: no award
    \end{itemize}

    \textbf{Chocolate products (providing origin):} 
    \begin{itemize}
        \item \texttt{Bohol}: Bohol island cacao
        \item \texttt{Davao}: Davao region cacao
        \item \texttt{Improved Bicol}: Bicol region cacao
    \end{itemize}
    \end{quote}

    \item \textbf{Additional Information in Few-shot Setting:}
    \begin{quote}
    ``To help with the estimation task, we provide some examples of a respondent’s decision as guidance for the model.

    \textbf{Examples (totally 6 examples):} 
    \begin{itemize}
        \item \texttt{0}: A person made a decision of: \{three WTPs correspond to three chocolate products\}
        \item \texttt{1}: A person made a decision of: \{three WTPs correspond to three chocolate products\}
        \item ...
        \item \texttt{5}: A person made a decision of: \{three WTPs correspond to three chocolate products\}
    \end{itemize}
        
    \end{quote}

\end{enumerate}

For each of the two information settings, we test four ways of prompting an LLM to generate a distribution.

\begin{enumerate}
    \item \textbf{Sampling:} 
    \begin{quote}
        ``Based on the information provided, please finish the task below. Suppose you hold an endowment chocolate (regular Bicol) worth 44 PHP. 
        
        \textbf{Task:} For each target product, report the premium (additional PHP over 44) you'd be willing to pay to exchange for it.
        
        \textbf{INSTRUCTIONS:}
        \begin{enumerate}
            \item First, explain your reasoning. Consider the chocolate item descriptions (and examples and persona if we provided). 
            \item Second, provide the premium (additional PHP over 44) you'd be willing to pay to exchange for each target product, and each premium value must be between 0 and 100. Use non-negative numbers; if you would not exchange for that product, use 0.
        \end{enumerate}
        \textbf{Output Format:} Reasoning: [...]
Final Answer: [\{"Bohol": $X$, "Davao": $Y$, "ImprovedBicol": $Z$\}]

\end{quote}

\item \textbf{Persona-sampling:} Same as sampling, but we provide the following additional instructions.

\begin{quote}
    ``To help with the estimation task, we provide a persona description of the one you should pretend to be. 

    So, pretend you are \{persona description\}.

    \textbf{Persona description (outside the distribution): } 
    {\footnotesize
    \begin{itemize}
  \setlength{\itemsep}{0pt}
  \setlength{\parskip}{0pt}
  \setlength{\parsep}{0pt}
  \item \texttt {Gender and age group}, age is divided into four bins (18–24, 25–34, 35–44, 45+).
  \item \texttt {Income}, classified as either "low" or "middle-to-high".
  \item \texttt {Main shopper status}, whether the user acts as the main shopper for the household (yes/no).
  \item \texttt{Consumption frequency of Tablea}, categorized into low, middle, or high levels.
  \item \texttt{Origin sensitivity}, the importance placed on product origin, rated on a 5-point scale ranging from "not important" to "very important". 
  \item \texttt{Chocolate preference}, intensity of chocolate preference (derived from dark chocolate data), classified as low, middle, or high.
  \item \texttt{Award influence}, influence of industry awards, classified into low, moderate, or high.
    \end{itemize}}
        
          \end{quote}

\item \textbf{Batch-generation:}
\begin{quote}
        ``Based on the information provided, please finish the task below. Suppose you hold an endowment chocolate (regular Bicol) worth 44 PHP. 
        
        \textbf{Task:} For each target product, provide 25 independent premiums (additional PHP over 44) you'd be willing to pay to exchange for it.
        
        \textbf{INSTRUCTIONS:}
        \begin{enumerate}
            \item First, explain your reasoning. Consider the sushi item descriptions (and examples if we provided).
            \item Second, provide 25 premiums (additional PHP over 44) you'd be willing to pay to exchange for each target product, and each premium value must be between 0 and 100. Use non-negative numbers; if you would not exchange for that product, use 0.
        \end{enumerate}
        \textbf{Output Format:} Reasoning: [...]
Final Answer: [\{"Bohol": $X_1$, "Davao": $Y_1$, "ImprovedBicol": $Z_1$\},\{"Bohol": $X_2$, "Davao": $Y_2$, "ImprovedBicol": $Z_2$\},...,\{"Bohol": $X_{25}$, "Davao": $Y_{25}$, "ImprovedBicol": $Z_{25}$\}]
\end{quote}

\item \textbf{Description:}
    \begin{quote}
        ``Based on the information provided, please finish the task below. Suppose you hold an endowment chocolate (regular Bicol) worth 44 PHP. 
        
        \textbf{Task:} For each target product, provide a discrete distribution over the premium (additional PHP over 44) you would be willing to pay to exchange for it.
        
        \textbf{INSTRUCTIONS:}
        \begin{enumerate}
            \item First, explain your reasoning. Consider the sushi item descriptions (and examples if we provided).
            \item Second, for each product, output exactly 5 most likely premium values. The 5 values must include 0 (meaning you would not exchange), and each premium value must be between 0 and 100.  Also, output the probability for each of the 5 values. Probabilities must be non-negative and sum to 1 for each product. Please avoid identical distributions across the three products unless strongly justified.
            
        \end{enumerate}
        \textbf{Output Format:} Reasoning: [...] Final Answer: [
\begin{verbatim}
{
  "premium_support": {
    "Bohol": [v1, v2, ..., v5],
    "Davao": [v1, v2, ..., v5],
    "ImprovedBicol": [v1, v2, ..., v5]
  },
  "probabilities": {
    "Bohol": [p1, p2, ..., p5],
    "Davao": [p1, p2, ..., p5],
    "ImprovedBicol": [p1, p2, ..., p5]
  }
}
\end{verbatim} ]
\end{quote}

\end{enumerate}

\clearpage
\section{Newsvendor problem: Prompts} 
\label{app:promptNews}

\begin{quote}
``You are an expert in inventory demand forecasting. Use the provided reference items to estimate statistics for the Target Item.

\textbf{Context (100 Reference Items with known Demand Statistics):}
\begin{verbatim}
[Reference Item 1]
Features:
- prod_name: [Value]
- product_type_name: [Value]
- graphical_appearance_name: [Value]
- colour_group_name: [Value]
- department_name: [Value]
- index_name: [Value]
- section_name: [Value]
- garment_group_name: [Value]
- detail_desc: [Value]

KNOWN STATISTICS -> Mean: [Value], Std: [Value]
------------------------------
... [Items 2-99] ...
------------------------------

Target Item Task:
Please estimate the demand distribution for:
- Product Name: [Target Item Name]
- Features:
    - prod_name: [Value]
    - product_type_name: [Value]
    - graphical_appearance_name: [Value]
    - colour_group_name: [Value]
    - department_name: [Value]
    - index_name: [Value]
    - section_name: [Value]
    - garment_group_name: [Value]
    - detail_desc: [Value]
\end{verbatim}

\textbf{Task:} Estimate the underlying Normal distribution parameters for the Target Item.

\textbf{STRICT OUTPUT FORMAT:} \\
You must start your response with exactly this line: \\
PREDICTION $\rightarrow$ Mean: [Value], Std: [Value]

Then, on a new line, provide your: \\
REASONING $\rightarrow$ [Your detailed explanation here]''
\end{quote}

\clearpage

\end{document}